\documentclass[letterpaper]{article}

\usepackage{times}
\usepackage{helvet}
\usepackage{courier}
\usepackage{url}
\usepackage{graphicx}
\usepackage[utf8]{inputenc} 
\usepackage[T1]{fontenc}    
\usepackage{booktabs}       
\usepackage{amsfonts}       
\usepackage{nicefrac}       
\usepackage{microtype}      
\usepackage{tikz}
\usetikzlibrary{shapes.misc, positioning}
\usepackage{thmtools,thm-restate}
\usepackage{parskip}
\usepackage{authblk}

\usepackage{floatrow}
\newfloatcommand{capbtabbox}{table}[][\FBwidth]

\usepackage{macros}
\usepackage{basic}

\newcommand{\AvgGainOverGold}[0]{20.2}
\newcommand{\AvgGainOverMV}[0]{6.8}
\newcommand{\AvgGainOverDP}[0]{4.1}

\newcommand{\AvgEMOverLM}[0]{3.4}

\newcommand{\AvgUnipolarBoost}[0]{2.8}

\newcommand{\AvgNumLFsPerTask}[0]{13}
\newcommand{\AvgLinesCodePerLF}[0]{4}

\title{Training Complex Models with Multi-Task Weak Supervision}

\author[$\dagger$]{Alexander~Ratner}
\author[$\dagger$]{Braden~Hancock}
\author[$\dagger$]{Jared~Dunnmon}

\author[$\dagger$]{Frederic~Sala}
\author[$\dagger$]{Shreyash~Pandey}
\author[$\dagger$]{Christopher~R{\'e}}

\affil[$\dagger$]{Department of Computer Science, Stanford University}
\affil[ ]{\footnotesize{\texttt{\{ajratner, bradenjh, jdunnmon, fredsala, shreyash, chrismre\}@stanford.edu}}}

\begin{document}

\maketitle

\newcommand{\figpathA}{fig/LF_example2b_wide.png}
\newcommand{\figwidthA}{\textwidth}
\newcommand{\figpathB}{fig/ontonotes_scaleup_plot_log.pdf}
\newcommand{\figwidthB}{0.5\linewidth}

\newcommand{\versionswitch}[2]{#2}

\begin{abstract}

As machine learning models continue to increase in complexity, collecting large hand-labeled training sets has become one of the biggest roadblocks in practice.
Instead, weaker forms of supervision that provide noisier but cheaper labels are often used.
However, these weak supervision sources have diverse and unknown accuracies, may output correlated labels, and may label different tasks or apply at different levels of granularity.
We propose a framework for integrating and modeling such weak supervision sources by viewing them as labeling different related sub-tasks of a problem, which we refer to as the \textit{multi-task weak supervision} setting.
We show that by solving a matrix completion-style problem, we can recover the accuracies of these \textit{multi-task} sources given their dependency structure, but without any labeled data, leading to higher-quality supervision for training an end model.
Theoretically, we show that the generalization error of models trained with this approach improves with the number of \textit{unlabeled} data points, and characterize the scaling with respect to the task and dependency structures.
On three fine-grained classification problems, we show that our approach leads to average gains of $\AvgGainOverGold$ points in accuracy over a traditional supervised approach, $\AvgGainOverMV$ points over a majority vote baseline, and $\AvgGainOverDP$ points over a previously proposed weak supervision method that models tasks separately.

\end{abstract}

\section{Introduction}
\label{sec:intro}
One of the greatest roadblocks to using modern machine learning models is collecting hand-labeled training data at the massive scale they require.
In real-world settings where domain expertise is needed and modeling goals change frequently, hand-labeling training sets is prohibitively slow, expensive, and static.
For these reasons, practitioners are increasingly turning to weak supervision techniques wherein noisier, often programmatically-generated labels are used instead.
Common \textit{weak supervision sources} include external knowledge bases~\cite{mintz2009distant,zhang:cacm17,craven:ismb99,takamatsu:acl12}, heuristic patterns~\cite{gupta2014improved,ratner2018snorkel}, feature annotations~\cite{mann2010generalized,zaidan:emnlp08}, and noisy crowd labels~\cite{karger2011iterative,dawid1979maximum}. The use of these sources has led to state-of-the-art results in a range of domains~\cite{zhang:cacm17,xiao2015learning}.
A theme of weak supervision is that using the full diversity of available sources is critical to training high-quality models~\cite{ratner2018snorkel,zhang:cacm17}.

The key technical difficulty of weak supervision is determining how to combine the labels of multiple sources that have different, unknown accuracies, may be correlated, and may label at different levels of granularity.
In our experience with users in academia and industry, the complexity of real world weak supervision sources makes this integration phase the key time sink and stumbling block.
For example, if we are training a model to classify entities in text, we may have one available source of high-quality but coarse-grained labels (e.g. ``Person'' vs. ``Organization'') and one source that provides lower-quality but finer-grained labels (e.g. ``Doctor'' vs. ``Lawyer''); moreover, these sources might be correlated due to some shared component or data source~\cite{bach2017learning,varma2017inferring}.
Handling such diversity requires addressing a core technical challenge: estimating the unknown accuracies of multi-granular and potentially correlated supervision sources without any labeled data.

To overcome this challenge, we propose \systemx, a framework for modeling and integrating weak supervision sources with different unknown accuracies, correlations, and granularities.
In \systemx, we view each source as labeling one of several related sub-tasks of a problem---we refer to this as the \textit{multi-task weak supervision} setting.
We then show that given the dependency structure of the sources, we can use their observed agreement and disagreement rates to recover their unknown accuracies.
Moreover, we exploit the relationship structure between tasks to observe additional cross-task agreements and disagreements, effectively providing extra signal from which to learn.
In contrast to previous approaches based on sampling from the posterior of a graphical model directly~\cite{ratner2016data,bach2017learning}, we develop a simple and scalable matrix completion-style algorithm, which we are able to analyze by applying strong matrix concentration bounds~\cite{tropp2015introduction}.
We use this algorithm to learn and model the accuracies of diverse weak supervision sources, and then combine their labels to produce training data that can be used to supervise arbitrary models, including increasingly popular multi-task learning models~\cite{Caruana93multitasklearning,DBLP:journals/corr/Ruder17a}.

Compared to previous methods which only handled the single-task setting~\cite{ratner2016data,ratner2018snorkel}, and generally considered conditionally-independent sources~\cite{anandkumar2014tensor,dawid1979maximum}, we demonstrate that our multi-task aware approach leads to average gains of $4.1$ points in accuracy in our experiments, and has at least three additional benefits.
First, many dependency structures between weak supervision sources may lead to non-identifiable models of their accuracies, where a unique solution cannot be recovered.
We provide a compiler-like check to establish identifiability---i.e. the existence of a unique set of source accuracies---for arbitrary dependency structures, without resorting to the standard assumption of non-adversarial sources~\cite{dawid1979maximum}, alerting users to this potential stumbling block that we have observed in practice.
Next, we provide sample complexity bounds that characterize the benefit of adding additional unlabeled data and the scaling with respect to the user-specified task and dependency structure.
While previous approaches required thousands of sources to give non-vacuous bounds, we capture regimes with small numbers of sources, better reflecting the real-world uses of weak supervision we have observed.
Finally, we are able to solve our proposed problem directly with SGD, leading to over $100\times$ faster runtimes compared to prior Gibbs-sampling based approaches~\cite{ratner2016data,platanios2017estimating}, and enabling simple implementation using libraries like PyTorch.

We validate our framework on three fine-grained classification tasks in named entity recognition, relation extraction, and medical document classification, for which we have diverse weak supervision sources at multiple levels of granularity.
We show that by modeling them as labeling hierarchically-related sub-tasks and utilizing unlabeled data, we can get an average improvement of $\AvgGainOverGold$ points in accuracy over a traditional supervised approach, $\AvgGainOverMV$ points over a basic majority voting weak supervision baseline, and $\AvgGainOverDP$ points over data programming~\cite{ratner2016data}, an existing weak supervision approach in the literature that is not multi-task-aware.
We also extend our framework to handle unipolar sources that only label one class, a critical aspect of weak supervision in practice that leads to an average $\AvgUnipolarBoost$ point contribution to our gains over majority vote.
From a practical standpoint, we argue that our framework represents an efficient way for practitioners to supervise modern machine learning models, including new multi-task variants, for complex tasks by opportunistically using the diverse weak supervision sources available to them.
To further validate this, we have released an open-source implementation of our framework.\footnote{\url{github.com/HazyResearch/metal}}

\begin{figure*}
	\centering
	\includegraphics[width=5in]{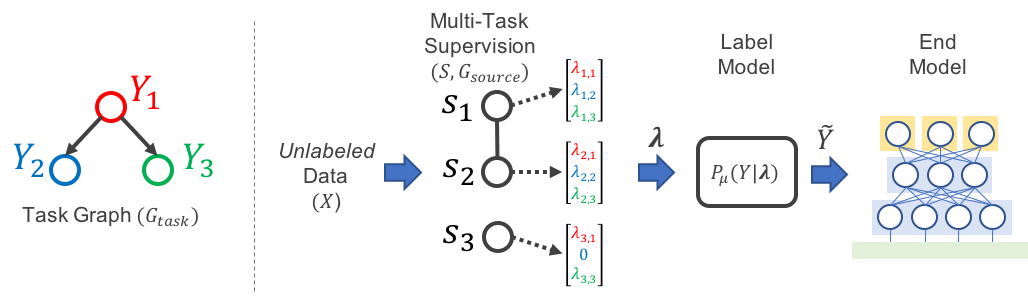}
	\caption{
		A schematic of the \systemx pipeline.
		To generate training data for an \textit{end model}, such as a multi-task model as in our experiments, the user inputs a \textit{task graph} $G_{\text{task}}$ defining the relationships between \textit{task labels} $Y_1,...,Y_t$; a set of \textit{unlabeled} data points $X$; a set of \textit{multi-task weak supervision sources} $s_i$ which each output a vector $\lf_i$ of task labels for $X$; and the dependency structure between these sources, $G_{\text{source}}$.
		We train a \textit{label model} to learn the accuracies of the sources, outputting a vector of probabilistic training labels $\tilde{\y}$ for training the end model.
	}
	\label{fig:ws_pipeline}
\end{figure*}

\section{Related Work}
\label{sec:related_work}

Our work builds on and extends various settings studied in machine learning.

\textit{Weak Supervision:}
We draw motivation from recent work which models and integrates weak supervision using generative models~\cite{ratner2016data,ratner2018snorkel,bach2017learning} and other methods~\cite{guan2017said,khetan2017learning}.
These approaches, however, do not handle multi-granularity or multi-task weak supervision, require expensive sampling-based techniques that may lead to non-identifiable solutions, and leave room for sharper theoretical characterization of weak supervision scaling properties. 
More generally, our work is motivated by a wide range of specific weak supervision techniques, which include traditional distant supervision approaches~\cite{mintz2009distant,craven:ismb99,zhang:cacm17,hoffmann:acl11,takamatsu:acl12}, co-training methods~\cite{blum1998combining}, pattern-based supervision~\cite{gupta2014improved,zhang:cacm17}, and feature-annotation techniques~\cite{mann2010generalized,zaidan:emnlp08,liang:icml09}.

\textit{Crowdsourcing:}
Our approach also has connections to the crowdsourcing literature~\cite{karger2011iterative,dawid1979maximum}, and in particular to spectral and method of moments-based approaches~\cite{zhang2014spectral,dalvi:www13,Ghosh:2011:MMC:1993574.1993599,anandkumar2014tensor}.
In contrast, the goal of our work is to support and explore settings not covered by crowdsourcing work, such as sources with correlated outputs, the proposed multi-task supervision setting, and regimes wherein a small number of labelers (weak supervision sources) each label a large number of items (data points).
Moreover, we theoretically characterize the generalization performance of an end model trained with the weakly labeled data.

\textit{Multi-Task Learning:}
Our proposed approach is motivated by recent progress on multi-task learning models~\cite{Caruana93multitasklearning,DBLP:journals/corr/Ruder17a,sogaard2016deep}, in particular their need for multiple large hand-labeled training datasets.
We note that the focus of our paper is on generating supervision for these models, not on the particular multi-task learning model being trained, which we seek to control for by fixing a simple architecture in our experiments.

Our work is also related to recent techniques for estimating classifier accuracies without labeled data in the presence of structural constraints~\cite{platanios2017estimating}.
We use matrix structure estimation~\cite{loh2012structure} and concentration bounds~\cite{tropp2015introduction} for our core results.

\section{Programming Machine Learning with Weak Supervision}
\label{sec:setup}

\begin{figure}
	\centering
	\includegraphics[width=\figwidthA]{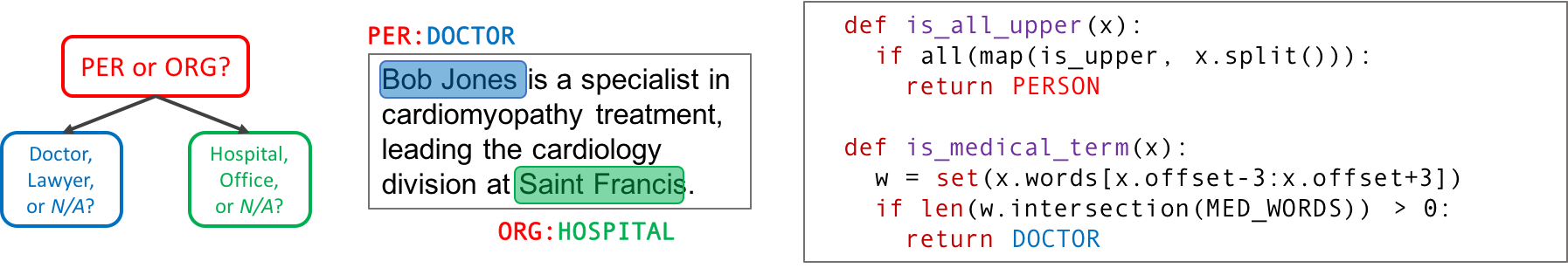}
	\caption{
		An example fine-grained entity classification problem, where weak supervision sources label three sub-tasks of different granularities: (i) \texttt{Person} vs. \texttt{Organization}, (ii) \texttt{Doctor} vs. \texttt{Lawyer} (or \textit{\texttt{N/A}}), (iii) \texttt{Hospital} vs. \texttt{Office} (or \textit{\texttt{N/A}}).
		The example weak supervision sources use a pattern heuristic and dictionary lookup respectively.
	}
	\label{fig:lf_example}
\end{figure}

As modern machine learning models become both more complex and more performant on a range of tasks, developers increasingly interact with them by programmatically generating noisier or \textit{weak} supervision.
These approaches of effectively \textit{programming} machine learning models have recently been formalized by the following pipeline~\cite{ratner2016data,ratner2018snorkel}:
First, users provide one or more \textit{weak supervision sources}, which are applied to unlabeled data to generate a set of noisy labels.
These labels may overlap and conflict; we model and combine them via a \textit{label model} in order to produce a final set of training labels.
These labels are then used to train some discriminative model, which we refer to as the \textit{end model}.
This programmatic weak supervision approach can utilize sources ranging from heuristic rules to other models, and in this way can also be viewed as a pragmatic and flexible form of multi-source \textit{transfer learning}.

In our experiences with users from science and industry, we have found it critical to utilize all available sources of weak supervision for complex modeling problems, including ones which label at multiple levels of \textit{granularity}.
However, this diverse, multi-granular weak supervision does not easily fit into existing paradigms.
We propose a formulation where each weak supervision source labels some sub-task of a problem, which we refer to as the \textit{multi-task weak supervision} setting.
We consider an example:
\begin{example}
	\label{ex:example-1}
	A developer wants to train a fine-grained Named Entity Recognition (NER) model to classify mentions of entities in the news (Figure~\ref{fig:lf_example}).
	She has a multitude of available weak supervision sources which she believes have relevant signal for her problem---for example, pattern matchers, dictionaries, and pre-trained generic NER taggers.
	However, it is unclear how to properly use and combine them: some of them label phrases coarsely as \texttt{PERSON} versus \texttt{ORGANIZATION}, while others classify specific fine-grained types of people or organizations, with a range of unknown accuracies.
	In our framework, she can represent them as labeling tasks of different granularities---e.g. $Y_1=\{\textrm{\texttt{Person}}, \textrm{\texttt{Org}}\}$, $Y_2=\{\textrm{\texttt{Doctor}}, \textrm{\texttt{Lawyer}, \texttt{N/A}}\}$, $Y_3=\{\textrm{\texttt{Hospital}}, \textrm{\texttt{Office}, \texttt{N/A}}\}$, where the label $\texttt{N/A}$ applies, for example, when the type-of-person task is applied to an organization.
\end{example}

In our proposed multi-task supervision setting, the user specifies a set of structurally-related \textit{tasks}, and then provides a set of \textit{weak supervision sources} which are user-defined functions that either label each data point or abstain for each task, and may have some user-specified dependency structure.
These sources can be arbitrary black-box functions, and can thus subsume a range of weak supervision approaches relevant to both text and other data modalities, including use of pattern-based heuristics, distant supervision~\cite{mintz2009distant}, crowd labels, other weak or biased classifiers, declarative rules over unsupervised feature extractors~\cite{varma2017inferring}, and more.
Our goal is to estimate the unknown accuracies of these sources, combine their outputs, and use the resulting labels to train an end model.

\section{Modeling Multi-Task Weak Supervision}
\label{sec:label_model}

The core technical challenge of the \textit{multi-task weak supervision} setting is recovering the unknown \textit{accuracies} of weak supervision sources given their dependency structure and a schema of the tasks they label, but without any ground-truth labeled data.
We define a new algorithm for recovering the accuracies in this setting using a matrix completion-style optimization objective.
We establish conditions under which the resulting estimator returns a unique solution.
We then analyze the sample complexity of our estimator, characterizing its scaling with respect to the amount of \textit{unlabeled data}, as well as the task schema and dependency structure, and show how the estimation error affects the generalization performance of the end model we aim to train.
Finally, we highlight how our approach handles abstentions and \textit{unipolar} sources, two critical scenarios in the weak supervision setting.

\subsection{A Multi-Task Weak Supervision Estimator}

\paragraph*{Problem Setup}
Let $X \in \mathcal{X}$ be a data point and $\y = [Y_1, Y_2, \ldots, Y_t]^T$ be a vector of categorical \textit{task labels}, $Y_i \in \{ 1,\ldots,k_i \}$, corresponding to $t$ tasks, where $(X,\y)$ is drawn i.i.d. from a distribution $\mathcal{D}$\versionswitch{.\footnote{The variables we introduce throughout this section are summarized in a glossary in the Appendix, which can be accessed at \url{https://arxiv.org/abs/1810.02840}.}}{ (for a glossary of all variables used, see Appendix~\ref{appendix:glossary}).}

The user provides a specification of how these tasks relate to each other; we denote this schema as the \textit{task structure} $G_{\text{task}}$.
The task structure expresses logical relationships between tasks, defining a \textit{feasible set} of label vectors $\mathcal{Y}$, such that $\y \in \mathcal{Y}$.
For example, Figure~\ref{fig:lf_example} illustrates a hierarchical task structure over three tasks of different granularities pertaining to a fine-grained entity classification problem.
Here, the tasks are related by logical subsumption relationships: for example, if $Y_2 = \texttt{DOCTOR}$, this implies that $Y_1 = \texttt{PERSON}$, and that $Y_3 = \textit{\texttt{N/A}}$, since the task label $Y_3$ concerns types of organizations, which is inapplicable to persons.
Thus, in this task structure, $\y = [\texttt{PERSON}, \texttt{DOCTOR}, \textit{\texttt{N/A}}]^T$ is in $\mathcal{Y}$ while $\y = [\texttt{PERSON}, \textit{\texttt{N/A}}, \texttt{HOSPITAL}]^T$ is not.
While task structures are often simple to define, as in the previous example, or are explicitly defined by existing resources---such as ontologies or graphs---we note that if no task structure is provided, our approach becomes equivalent to modeling the $t$ tasks separately, a baseline we consider in the experiments.

In our setting, rather than observing the true label $\y$, we have access to $m$ \textit{multi-task weak supervision} sources $s_i \in S$ which emit label vectors $\lf_i$ that contain labels for some subset of the $t$ tasks.
Let $0$ denote a null or abstaining label, and let the \textit{coverage set} $\tau_i \subseteq \{1, \ldots, t\}$ be the fixed set of tasks for which the $i$th source emits non-zero labels, such that $\lf_i \in \mathcal{Y}_{\tau_{i}}$.
For convenience, we let $\tau_0 = \{1, \ldots, t\}$ so that $\mathcal{Y}_{\tau_0} = \mathcal{Y}$.
For example, a source from our previous example might have a coverage set $\tau_i = \{1,3\}$, emitting coarse-grained labels such as $\lf_i = [\texttt{PERSON}, 0, \textit{\texttt{N/A}}]^T$.
Note that sources often label multiple tasks implicitly due to the constraints of the task structure; for example, a source that labels types of people ($Y_2$) also implicitly labels people vs. organizations ($Y_1 = \texttt{PERSON}$), and types of organizations (as $Y_3 = \textit{\texttt{N/A}}$).
Thus sources tailored to different tasks still have agreements and disagreements; we use this additional \textit{cross-task} signal in our approach.

\versionswitch{
	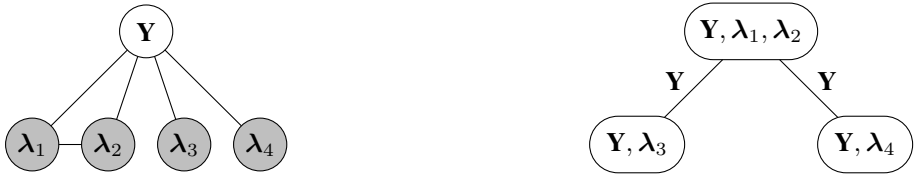
\begin{figure}
		\begin{subfigure}{.5\linewidth}
			\centering
			\begin{tikzpicture}[every node/.style={inner sep=0,outer sep=0}]
			
			\draw (0,1.25) node[draw,circle,fill=white,minimum size=0.6cm](y) {$\y$};
			
			\draw (-1.35,0) node[draw,fill=lightgray,circle,minimum size=0.6cm](l1x) {$\lf_1$};
			\draw (-0.45,0) node[draw,fill=lightgray,circle,minimum size=0.6cm](l2x) {$\lf_2$};
			\draw (0.45,0) node[draw,fill=lightgray,circle,minimum size=0.6cm](l3x) {$\lf_3$};
			\draw (1.35,0) node[draw,fill=lightgray,circle,minimum size=0.6cm](l4x) {$\lf_4$};
	
			\draw (y) -- (l1x);
			\draw (y) -- (l2x);
			\draw (l1x) -- (l2x);
			\draw (y) -- (l3x);
			\draw (y) -- (l4x);
			
			\end{tikzpicture}
			\label{fig:g-source}
		\end{subfigure}%
		\begin{subfigure}{.5\linewidth}
			\centering
			\begin{tikzpicture}[every node/.style={inner sep=0,outer sep=0}]
			
			\draw (0,1.25) node [
				draw,
				rounded rectangle,
				fill=white,
				minimum height=0.75cm,
				minimum width=1.75cm
			] (jtn1) {$\y, \lf_1, \lf_2$};
			\draw (-1,0) node [
				draw,
				rounded rectangle,
				fill=white,
				minimum height=0.75cm,
				minimum width=1.25cm
			] (jtn2) {$\y, \lf_3$};
			\draw (1,0) node [
				draw,
				rounded rectangle,
				fill=white,
				minimum height=0.75cm,
				minimum width=1.25cm
			] (jtn3) {$\y, \lf_4$};
	
			\draw (jtn1) -- (jtn2);
			\draw (jtn1) -- (jtn3);
	
			\draw (1,0.75) node [] (ss12) {$\y$};
			\draw (-1,0.75) node [] (ss13) {$\y$};
			
			\end{tikzpicture}
			\label{fig:g-source-jt}
		\end{subfigure}%
		\caption{
			An example of a weak supervision source dependency graph $G_{\text{source}}$ (left) and its junction tree representation (right), where $\y$ is a vector-valued random variable with a feasible set of values, $\y \in \mathcal{Y}$.
			Here, the output of sources 1 and 2 are modeled as dependent conditioned on $\y$.
			This results in a junction tree with singleton separator sets, $\y$.
			Here, the observable cliques are $O = \{\lf_1,\lf_2,\lf_3,\lf_4,\{\lf_1,\lf_2\}\} \subset \mathcal{C}$.
		}
		\label{fig:g-source-example}
	\end{figure}
}{
\begin{figure}
	\begin{subfigure}{.5\textwidth}
		\centering
		\begin{tikzpicture}[every node/.style={inner sep=0,outer sep=0}]
		
		\draw (0,1.5) node[draw,circle,fill=white,minimum size=0.7cm](y) {$\y$};
		
		\draw (-1.5,0) node[draw,fill=lightgray,circle,minimum size=0.7cm](l1x) {$\lf_1$};
		\draw (-0.5,0) node[draw,fill=lightgray,circle,minimum size=0.7cm](l2x) {$\lf_2$};
		\draw (0.5,0) node[draw,fill=lightgray,circle,minimum size=0.7cm](l3x) {$\lf_3$};
		\draw (1.5,0) node[draw,fill=lightgray,circle,minimum size=0.7cm](l4x) {$\lf_4$};

		\draw (y) -- (l1x);
		\draw (y) -- (l2x);
		\draw (l1x) -- (l2x);
		\draw (y) -- (l3x);
		\draw (y) -- (l4x);
		
		\end{tikzpicture}
		\label{fig:g-source}
	\end{subfigure}%
	\begin{subfigure}{.5\textwidth}
		\centering
		\begin{tikzpicture}[every node/.style={inner sep=0,outer sep=0}]
		
		\draw (0,1.5) node [
			draw,
			rounded rectangle,
			fill=white,
			minimum height=0.75cm,
			minimum width=1.75cm
		] (jtn1) {$\y, \lf_1, \lf_2$};
		\draw (-1.5,0) node [
			draw,
			rounded rectangle,
			fill=white,
			minimum height=0.75cm,
			minimum width=1.25cm
		] (jtn2) {$\y, \lf_3$};
		\draw (1.5,0) node [
			draw,
			rounded rectangle,
			fill=white,
			minimum height=0.75cm,
			minimum width=1.25cm
		] (jtn3) {$\y, \lf_4$};

		\draw (jtn1) -- (jtn2);
		\draw (jtn1) -- (jtn3);

		\draw (1,0.85) node [] (ss12) {$\y$};
		\draw (-1,0.85) node [] (ss13) {$\y$};
		
		\end{tikzpicture}
		\label{fig:g-source-jt}
	\end{subfigure}%
	\caption{
		An example of a weak supervision source dependency graph $G_{\text{source}}$ (left) and its junction tree representation (right), where $\y$ is a vector-valued random variable with a feasible set of values, $\y \in \mathcal{Y}$.
		Here, the output of sources 1 and 2 are modeled as dependent conditioned on $\y$.
		This results in a junction tree with singleton separator sets, $\y$.
		Here, the observable cliques are $O = \{\lf_1,\lf_2,\lf_3,\lf_4,\{\lf_1,\lf_2\}\} \subset \mathcal{C}$.
	}
	\label{fig:g-source-example}
\end{figure}
}

The user also provides the conditional dependency structure of the sources as a graph $G_{\text{source}} = (V,E)$, where $V = \{\y, \lf_1, \lf_2, \ldots, \lf_m\}$ (Figure~\ref{fig:g-source-example}).
Specifically, if $(\lf_i, \lf_j)$ is not an edge in $G_{\text{source}}$, this means that $\lf_i$ is independent of $\lf_j$ conditioned on $\y$ and the other source labels.
Note that if $G_{\text{source}}$ is unknown, it can be estimated using statistical techniques such as~\cite{bach2017learning}.
Importantly, we do not know anything about the strengths of the correlations in $G_{\text{source}}$, or the sources' accuracies.

Our overall goal is to apply the set of weak supervision sources $S = \{s_1, \ldots, s_m\}$ to an unlabeled dataset $\mathcal{X}_U$ consisting of $n$ data points, then use the resulting weakly-labeled training set to supervise an \textit{end model} $f_w : \mathcal{X} \mapsto \mathcal{Y}$ (Figure~\ref{fig:ws_pipeline}).
This weakly-labeled training set will contain overlapping and conflicting labels, from sources with unknown accuracies and correlations.
To handle this, we will learn a \textit{label model} $P_\mu(\y|\lf)$, parameterized by a vector of source correlations and accuracies $\mu$, which for each data point $X$ takes as input the noisy labels $\lf = \{\lf_1, \ldots, \lf_m\}$ and outputs a single probabilistic label vector $\tilde{\y}$.
Succinctly, given a user-provided tuple $(\mathcal{X}_U, S, G_{\text{source}}, G_{\text{task}})$, our key technical challenge is recovering the parameters $\mu$ without access to ground truth labels $\y$.

\paragraph*{Modeling Multi-Task Sources}
To learn a label model over multi-task sources, we introduce sufficient statistics over the random variables in $G_{\text{source}}$.
Let $\mathcal{C}$ be the set of cliques in $G_{\text{source}}$, and define an indicator random variable for the event of a clique $C \in \mathcal{C}$ taking on a set of values $y_C$:
\begin{align*}
	\psi(C,y_C)
	&=
	\ind{ \cap_{i \in C} V_i = (y_C)_i },
\end{align*}
where $(y_C)_i \in \mathcal{Y}_{\tau_i}$.
We define $\psi(C) \in \{0,1\}^{\prod_{i\in C} (|\mathcal{Y}_{\tau_i}|-1)}$ as the vector of indicator random variables for all combinations of all but one of the labels emitted by each variable in clique $C$---thereby defining a minimal set of statistics---and define $\psi(\textbf{C})$ accordingly for any set of cliques $\textbf{C} \subseteq \mathcal{C}$.
Then $\mu = \E{}{ \psi(\mathcal{C}) }$ is the vector of sufficient statistics for the label model we want to learn.

We work with two simplifying conditions in this section.
First, we consider the setting where $G_{\text{source}}$ is \textit{triangulated} and has a junction tree representation with singleton separator sets.
If this is not the case, edges can always be added to $G_{\text{source}}$ to make this setting hold; otherwise, we describe how our approach can directly handle non-singleton separator sets in \versionswitch{the Appendix}{Appendix~\ref{appendix:non-singleton-sep-sets}}.

Second, we use a simplified \textit{class-conditional} model of the noisy labeling process, where we learn one accuracy parameter for each label value $\lf_i$ that each source $s_i$ emits.
This is equivalent to assuming that a source may have a different accuracy on each different class, but that if it emits a certain label incorrectly, it does so uniformly over the different true labels $\y$.
This is a more expressive model than the commonly considered one, where each source is modeled by a single accuracy parameter, e.g. in~\cite{dawid1979maximum,ratner2016data}, and in particular allows us to capture the \textit{unipolar} setting considered later on.
\versionswitch{}{For further details, see Appendix~\ref{appendix:rank-one-reduction}.}

\paragraph*{Our Approach}
The chief technical difficulty in our problem is that we do not observe $\y$.
We overcome this by analyzing the covariance matrix of an observable subset of the cliques in $G_{\text{source}}$, leading to a matrix completion-style approach for recovering $\mu$.
We leverage two pieces of information: (i) the observability of \textit{part of} $\Cov{}{\psi(\mathcal{C})}$, and (ii) a result from \citep{loh2012structure} which states that the inverse covariance matrix $\Cov{}{\psi(\mathcal{C})}^{-1}$ is structured according to $G_{\text{source}}$, i.e., if there is no edge between $\lf_i$ and $\lf_j$ in $G_{\text{source}}$, then the corresponding entries are 0.

We start by considering two disjoint subsets of $\mathcal{C}$: the set of observable cliques, $O \subseteq \mathcal{C}$---i.e., those cliques not containing $\y$---and the separator set cliques of the junction tree, $\mathcal{S} \subseteq \mathcal{C}$. In the setting we consider in this section, $\mathcal{S} = \{\y\}$ (see Figure~\ref{fig:g-source-example}).
We can then write the covariance matrix of the indicator variables for $O \cup \mathcal{S}$, $\Cov{}{ \psi(O \cup \mathcal{S}) }$, in block form, similar to~\cite{chandrasekaran2010latent}, as:
\begin{align}
	\label{eqn:gen-cov}
	\Cov{}{ \psi(O \cup \mathcal{S}) }
	\equiv
	&~\Sigma
	=
	\begin{bmatrix}
		\Sigma_O & \Sigma_{O\mathcal{S}} \\
		\Sigma_{O\mathcal{S}}^T & \Sigma_\mathcal{S}
	\end{bmatrix}
\end{align}
and similarly define its inverse:
\begin{align}
	K
	&=
	\Sigma^{-1}
	=
	\begin{bmatrix}
		K_O & K_{O\mathcal{S}} \\
		K_{O\mathcal{S}}^T & K_\mathcal{S}
	\end{bmatrix}
\end{align}
Here, $\Sigma_O = \Cov{}{ \psi(O) } \in \mathbb{R}^{d_O \times d_O}$ is the observable block of $\Sigma$, where $d_O = \sum_{C \in O}\prod_{i \in C}(|\mathcal{Y}_{\tau_i}|-1)$.
Next, $\Sigma_{O\mathcal{S}} = \Cov{}{ \psi(O), \psi(\mathcal{S}) }$ is the unobserved block which is a function of $\mu$, the label model parameters that we wish to recover.
Finally, $\Sigma_\mathcal{S} = \Cov{}{ \psi(\mathcal{S}) } = \Cov{}{ \psi(\y) }$ is a function of the class balance $P(\y)$.

We make two observations about $\Sigma_\mathcal{S}$.
First, while the full form of $\Sigma_\mathcal{S}$ is the covariance of the $|\mathcal{Y}|-1$ indicator variables for each individual value of $\y$ but one, given our simplified class-conditional label model, we in fact only need a single indicator variable for $\y$ (see Appendix\versionswitch{}{ \ref{appendix:rank-one-reduction}}); thus, $\Sigma_\mathcal{S}$ is a scalar.
Second, $\Sigma_\mathcal{S}$ is a function of the class balance $P(\y)$, which we assume is either known, or has been estimated according to the unsupervised approach we detail in \versionswitch{the Appendix}{Appendix~\ref{appendix:recovering-class-balance}}.
Thus, given $\Sigma_O$ and  $\Sigma_\mathcal{S}$, our goal is to recover the vector $\Sigma_{O\mathcal{S}}$ from which we can recover $\mu$.
	
Applying the block matrix inversion lemma, we have:
\begin{align}
	\label{eqn:block-inv-cov-main}
	K_O
	&=
	\Sigma_O^{-1}
	+ c\Sigma_O^{-1}\Sigma_{O\mathcal{S}} \Sigma_{O\mathcal{S}}^T\Sigma_O^{-1},
\end{align}
where $c = \left( \Sigma_\mathcal{S} - \Sigma_{O\mathcal{S}}^T\Sigma_O^{-1}\Sigma_{O\mathcal{S}} \right)^{-1} \in \mathbb{R}^+$.
Let $z = \sqrt{c} \Sigma_O^{-1}\Sigma_{O\mathcal{S}}$; we can then express (\ref{eqn:block-inv-cov-main}) as:
\begin{align}
	\label{eqn:matrix-completion-form-main}
	K_O
	&=
	\Sigma_O^{-1} + zz^T
\end{align}
The right hand side of (\ref{eqn:matrix-completion-form-main}) consists of an empirically observable term, $\Sigma_O^{-1}$, and a rank-one term, $zz^T$, which we can solve for to directly recover $\mu$.
For the left hand side, we apply an extension of Corollary 1 from \citep{loh2012structure} (see Appendix\versionswitch{}{ \ref{appendix:model-estimation}}) to conclude that $K_O$ has graph-structured sparsity, i.e., it has zeros determined by the structure of dependencies between the sources in $G_{\text{source}}$.
This suggests an algorithmic approach of estimating $z$ as a matrix completion problem in order to recover an estimate of $\mu$ (Algorithm~\ref{alg:method}).
In more detail: let $\Omega$ be the set of indices $(i,j)$ where $(K_O)_{i,j} = 0$, determined by $G_{\text{source}}$, yielding a system of equations,
\begin{align}
	0
	&=
	(\Sigma_O^{-1})_{i,j}
	+ \left( zz^T \right)_{i,j} \text{ for } (i,j) \in \Omega,
	\label{eqn:constraints}
\end{align}
which is now a matrix completion problem.
Define $\norm{A}_{\Omega}$ as the Frobenius norm of $A$ with entries not in $\Omega$ set to zero; then we can rewrite (\ref{eqn:constraints}) as $\norm{ \Sigma_O^{-1} + zz^T }_\Omega = 0$.
We solve this equation to estimate $z$, and thereby recover $\Sigma_{O\mathcal{S}}$, from which we can directly recover the label model parameters $\mu$ algebraically.

\paragraph*{Checking for Identifiability}
A first question is: which dependency structures $G_{\text{source}}$ lead to unique solutions for $\mu$?
This question presents a stumbling block for users, who might attempt to use non-identifiable sets of correlated weak supervision sources.

We provide a simple, testable condition for identifiability.
Let $G_{\text{inv}}$ be the inverse graph of $G_{\text{source}}$; note that $\Omega$ is the edge set of $G_{\text{inv}}$ expanded to include all indicator random variables $\psi(\mathcal{C})$.
Then, let $M_{\Omega}$ be a matrix with dimensions $|\Omega| \times d_O$ such that each row in $M_{\Omega}$ corresponds to a pair $(i,j) \in \Omega$ with $1$'s in positions $i$ and $j$ and 0's elsewhere.

Taking the log of the squared entries of (\ref{eqn:constraints}), we get a system of linear equations $M_\Omega l = q_\Omega$,
where $l_i = \log(z_i^2)$ and $q_{(i,j)} = \log(((\Sigma_O^{-1})_{i,j})^2)$.
Assuming we can solve this system (which we can always ensure by adding sources; see Appendix), we can uniquely recover the $z_i^2$, meaning our model is identifiable \textit{up to sign}.

Given estimates of the $z_i^2$, we can see from (\ref{eqn:constraints}) that the sign of a single $z_i$ determines the sign of all other $z_j$ reachable from $z_i$ in $G_{\text{inv}}$.
Thus to ensure a unique solution, we only need to pick a sign for each connected component in  $G_{\text{inv}}$.
In the case where the sources are assumed to be independent, e.g.,~\cite{Dalvi:2013:ACB:2488388.2488414,zhang2014spectral,dawid1979maximum}, it
suffices to make the assumption that the sources are {\em on average} non-adversarial; i.e., select the sign of the $z_i$ that leads to higher average accuracies of the sources.
Even a single source that is conditionally independent from all the other sources will cause $G_{\text{inv}}$ to be fully connected, meaning we can use this symmetry breaking assumption in the majority of cases even with correlated sources.
Otherwise, a sufficient condition is the standard one of assuming non-adversarial sources, i.e. that all sources have greater than random accuracy.
\versionswitch{}{For further details, see Appendix~\ref{appendix:identifiability}.}

\paragraph*{Source Accuracy Estimation Algorithm}

\begin{figure*}
    \centering
    \includegraphics[width=\textwidth]{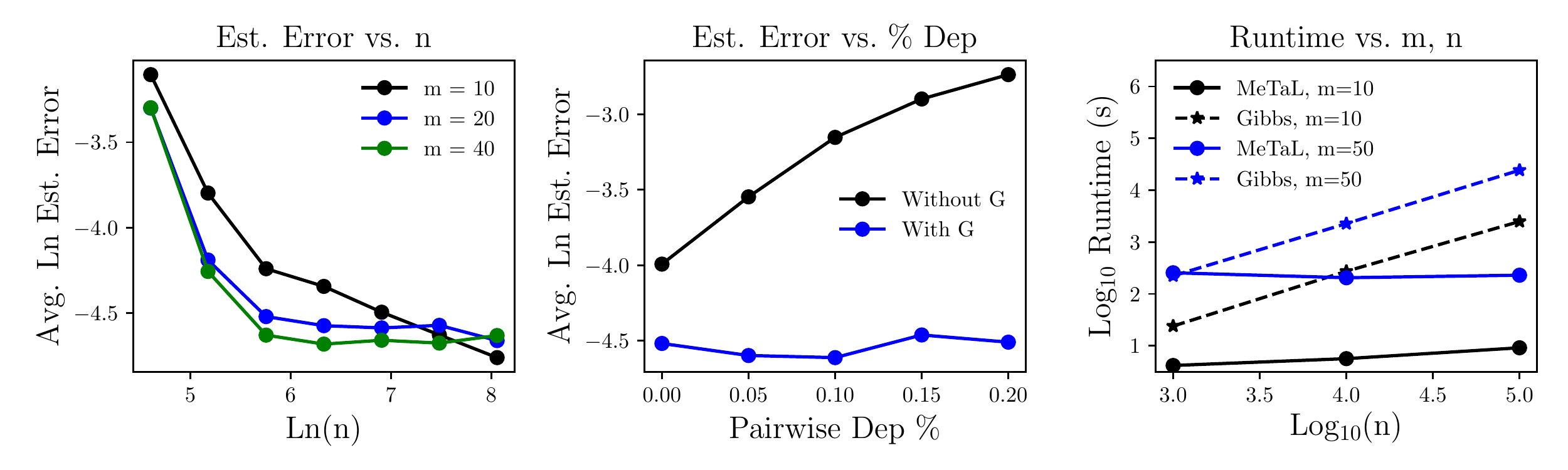}
    \caption{(Left) Estimation error $\norm{ \hat{\mu} - \mu^* }$ decreases with increasing $n$. (Middle) Given $G_{\text{source}}$, our model successfully recovers the source accuracies even with many pairwise dependencies among sources, where a naive conditionally-independent model fails. (Right) The runtime of \systemx~is independent of $n$ after an initial matrix multiply, and can thus be multiple orders of magnitude faster than Gibbs sampling-based approaches~\cite{ratner2016data}.}
    \label{fig:synthetic_fig}
\end{figure*}

Now that we know when a set of sources with correlation structure $G_{\text{source}}$
is identifiable, yielding a unique $z$, we can estimate the accuracies $\mu$ using Algorithm~\ref{alg:method}.
We also use the function ExpandTied, which is a simple algebraic expansion of tied parameters according to the simplified class-conditional model used in this section; see Appendix\versionswitch{}{ \ref{appendix:rank-one-reduction}} for details.
In Figure~\ref{fig:synthetic_fig}, we
plot the performance of our algorithm on synthetic data, showing its
scaling with the number of unlabeled data points $n$, the density of pairwise dependencies in $G_{\text{source}}$, and the runtime performance as compared to a prior Gibbs sampling-based approach. Next, we theoretically analyze
the scaling of the error $\norm{ \hat{\mu} - \mu^* }$.

\begin{algorithm}[tb]
	\caption{Source Accuracy Estimation for Multi-Task Weak Supervision}
   	\label{alg:method}
	\begin{algorithmic}
		\State \textbf{Input:}
			Observed labeling rates $\Ehat{\psi(O)}$ and covariance $\hat{\Sigma}_O$;
			class balance $\Ehat{\psi(\y)}$ and variance $\Sigma_\mathcal{S}$;
			correlation sparsity structure $\Omega$

		\State $\hat{z} \leftarrow \argmin{z}{ \norm{ \hat{\Sigma}_O^{-1} + zz^T }_\Omega}$

		\State $\hat{c} \leftarrow \Sigma_\mathcal{S}^{-1}(1 + \hat{z}^T\hat{\Sigma}_O\hat{z})$, $\hat{\Sigma}_{O\mathcal{S}} \leftarrow \hat{\Sigma}_O\hat{z} / \sqrt{\hat{c}}$
		\State $\hat{\mu}' \leftarrow \hat{\Sigma}_{O\mathcal{S}} + \Ehat{\psi(\y)} \Ehat{\psi(O)}$
			\\
		\Return $\text{ExpandTied}(\hat{\mu}')$
	\end{algorithmic}
\end{algorithm}

\subsection{Theoretical Analysis: Scaling with Diverse Multi-Task Supervision}
Our ultimate goal is to train an \textit{end model} using the source labels, denoised and combined by the label model $\hat{\mu}$ we have estimated.
We connect the generalization error of this end model to the estimation error of Algorithm~\ref{alg:method}, ultimately showing that the generalization error scales as $n^{-\frac12}$, where $n$ is the number of unlabeled data points.
This key result establishes the same asymptotic scaling as traditionally supervised learning methods, but with respect to \textit{unlabeled} data points.

Let $P_{\hat{\mu}}(\tilde{\y}~|~\lf)$ be the probabilistic label (i.e. distribution) predicted by our label model, given the source labels $\lf$ as input, which we compute using the estimated $\hat{\mu}$.
We then train an \textit{end} multi-task discriminative model $f_w : \mathcal{X} \mapsto \mathcal{Y}$ parameterized by $w$, by minimizing the expected loss with respect to the label model over $n$ unlabeled data points.
Let $l(w, X, \y) = \frac1t\sum_{s=1}^t l_t(w, X, \y_s)$ be a bounded multi-task loss function such that without loss of generality $l(w,X,\y) \leq 1$; then we minimize the empirical \textit{noise aware loss}:
\begin{align}
    \hat{w}
    &=
    \argmin{w}{
        \frac{1}{n} \sum_{i=1}^n
        \E{\tilde{\y} \sim P_{\hat{\mu}}(\cdot|\lf)}{ l(w, X_i, \tilde{\y}) }
    },
    \label{eqn:expected_loss}
\end{align}
and let $\tilde{w}$ be the $w$ that minimizes the true noise-aware loss.
This minimization can be performed by standard methods and is not the focus of our paper; let the solution $\hat{w}$ satisfy $\E{}{\|\hat{w} - \tilde{w}\|^2} \leq \gamma$.
We make several assumptions, following \cite{ratner2016data}: (1) that for some label model parameters $\mu^*$, sampling $(\lf, \y) \sim P_{\mu^*}(\cdot)$ is the same as sampling from the true distribution, $(\lf, \y) \sim \mathcal{D}$; and (2) that the task labels $Y_s$ are independent of the features of the end model given $\lf$ sampled from $P_{\mu^*}(\cdot)$, that is, the output of the optimal label model provides sufficient information to discern the true label.
Then we have the following result:
\begin{theorem}
	Let $\tilde{w}$ minimize the expected noise aware loss, using weak supervision source parameters $\hat{\mu}$ estimated with Algorithm~\ref{alg:method}.
	Let $\hat{w}$ minimize the empirical noise aware loss with $\E{}{\|\hat{w} - \tilde{w}\|^2} \leq \gamma$, $w^* = \min_{w} l(w, X,\y)$, and let the assumptions above hold. Then the generalization error is bounded by:
    \begin{align*}
        \E{}{ l(\hat{w},X, \y) -  l(w^*, X, \y) }
        &\leq
        \gamma + 4 |\mathcal{Y}| \norm{ \hat{\mu} - \mu^*}.
    \end{align*}
\end{theorem}
Thus, to control the generalization error, we must control $\norm{ \hat{\mu} - \mu^* }$, which
we do in Theorem~\ref{thm:mu_est}:

\versionswitch{
\begin{restatable}{theorem}{thmmuest}
\label{thm:mu_est}
Let $\hat{\mu}$ be an estimate of $\mu^*$ produced by Algorithm~\ref{alg:method} run over $n$ unlabeled data points.
Let $a := ( \frac{d_O}{\Sigma_{\mathcal{S}}} + (\frac{d_O}{\Sigma_{\mathcal{S}}})^2 \lambda_{\textrm{max}}(K_O) )^{\frac12}$ and $b := \frac{ \|\Sigma_O^{-1}\|^2}{(\Sigma_O^{-1})_{\min}}$.
Then, we have:
\begin{align*}
	&\E{}{\norm{ \hat{\mu} - \mu^*}}
	\leq 16(|\mathcal{Y}|-1)d_O^2 \sqrt{\frac{32 \pi}{n}} ab \sigma_{\max}(M_{\Omega}^{+})\\
	&\times
	\left( 3\sqrt{d_O}a \lambda_{\min}^{-1}(\Sigma_O) + 1 \right)
	\left( \kappa(\Sigma_O) + \lambda_{\min}^{-1}(\Sigma_O) \right).
\end{align*}
\end{restatable}
}{
\begin{restatable}{theorem}{thmmuest}
\label{thm:mu_est}
Let $\hat{\mu}$ be an estimate of $\mu^*$ produced by Algorithm~\ref{alg:method} run over $n$ unlabeled data points.
Let $a := \left( \frac{d_O}{\Sigma_{\mathcal{S}}} + \left(\frac{d_O}{\Sigma_{\mathcal{S}}}\right)^2 \lambda_{\textrm{max}}(K_O) \right)^{\frac12}$ and $b := \frac{ \|\Sigma_O^{-1}\|^2}{(\Sigma_O^{-1})_{\min}}$.
Then, we have:
\begin{align*}
	&\E{}{\norm{ \hat{\mu} - \mu^*}}
	\leq 16(r-1)d_O^2 \sqrt{\frac{32 \pi}{n}} ab \sigma_{\max}(M_{\Omega}^{+})
		\left( 3\sqrt{d_O}a \lambda_{\min}^{-1}(\Sigma_O) + 1 \right)
		\left( \kappa(\Sigma_O) + \lambda_{\min}^{-1}(\Sigma_O) \right).
\end{align*}
\end{restatable}
}
 
\paragraph{Interpreting the Bound} 
We briefly explain the key terms controlling the bound in Theorem~\ref{thm:mu_est}; more detail is found in \versionswitch{the Appendix}{Appendix~\ref{appendix:theoretical}}.
Our primary result is that the estimation error scales as $n^{-\frac12}$.
Next, $\sigma_{\max}(M_{\Omega}^+)$, the largest singular value of the pseudoinverse $M_{\Omega}^+$, has a deep connection to the density of the graph $G_{\text{inv}}$.
The smaller this quantity, the more information we have about $G_{\text{inv}}$, and the easier it is to estimate the accuracies. 
Next, $\lambda_{\min}(\Sigma_O)$, the smallest eigenvalue of the observed covariance matrix, reflects the conditioning of $\Sigma_O$; better conditioning yields easier estimation, and is roughly determined by how far away from random guessing the worst weak supervision source is, as well as how conditionally independent the sources are.
$\lambda_{\textrm{max}}(K_O)$, the largest eigenvalue of the upper-left block of the inverse covariance matrix, similarly reflects the overall conditioning of $\Sigma$.
Finally, $(\Sigma_O^{-1})_{\min}$, the smallest entry of the inverse observed matrix, reflects the smallest non-zero correlation between source accuracies; distinguishing between small correlations and independent sources requires more samples.

\subsection{Extensions: Abstentions \& Unipolar Sources}
\label{sec:extensions}
We briefly highlight two extensions handled by our approach which we have found empirically critical: handling \textit{abstentions}, and modeling \textit{unipolar} sources.

\textit{Handling Abstentions. }
One fundamental aspect of the weak supervision setting is that sources may abstain from labeling a data point entirely---that is, they may have incomplete and differing coverage~\cite{ratner2018snorkel,Dalvi:2013:ACB:2488388.2488414}.
We can easily deal with this case by extending the coverage ranges $\mathcal{Y}_{\tau_i}$ of the sources to include the vector of all zeros, $\vec{0}$, and we do so in the experiments.

\textit{Handling Unipolar Sources. }
Finally, we highlight the fact that our approach models \textit{class conditional} source accuracies, in particular motivated by the case we have frequently observed in practice of \textit{unipolar} weak supervision sources, i.e., sources that each only label a single class or abstain.
In practice, we find that users most commonly use such unipolar sources; for example, a common template for a heuristic-based weak supervision source over text is one that looks for a specific pattern, and if the pattern is present emits a specific label, else abstains. 
As compared to prior approaches that did not model class-conditional accuracies, e.g.~\cite{ratner2016data}, we show in our experiments that we can use our class-conditional modeling approach to yield an improvement of $\AvgUnipolarBoost$ points in accuracy.

\section{Experiments}
\label{sec:experiments}

\begin{table*}[t]
    \centering
    \begin{tabular}{lrrrr}
      \toprule
              & NER & RE & Doc & Average\\
      \midrule
      Gold (Dev)
        & 63.7 $\pm$ 2.1
        & 28.4 $\pm$ 2.3
        & 62.7 $\pm$ 4.5
        & 51.6 
        \\
      MV
        & 76.9 $\pm$ 2.6
        & 43.9 $\pm$ 2.6
        & 74.2 $\pm$ 1.2
        & 65.0
        \\
      DP \cite{ratner2016data}
        & 78.4 $\pm$ 1.2
        & 49.0 $\pm$ 2.7
        & 75.8 $\pm$ 0.9
        & 67.7
        \\
      \midrule
      \systemx
        & \textbf{82.2} $\pm$ 0.8
        & \textbf{56.7} $\pm$ 2.1
        & \textbf{76.6} $\pm$ 0.4
        & \textbf{71.8}
        \\
      \bottomrule
    \end{tabular}
    \caption{\textbf{Performance Comparison of Different Supervision Approaches.} We compare the micro accuracy (avg. over 10 trials) with 95\% confidence intervals of an end multi-task model trained using the training labels from the hand-labeled development set (Gold Dev), hierarchical majority vote (MV), data programming (DP), and our approach (\systemx).}
    \label{tab:results}
\end{table*}

We validate our approach on three fine-grained classification problems---entity classification, relation classification, and document classification---where weak supervision sources are available at both coarser and finer-grained levels (e.g. as in Figure~\ref{fig:lf_example}).
We evaluate the predictive accuracy of end models supervised with training data produced by several approaches, finding that our approach outperforms traditional hand-labeled supervision by 20.2 points, a baseline majority vote weak supervision approach by $\AvgGainOverMV$ points, and a prior weak supervision denoising approach~\cite{ratner2016data} that is not multi-task-aware by $\AvgGainOverDP$ points.

\paragraph*{Datasets}
\label{sec:data}
Each dataset consists of a large (3k-63k) amount of unlabeled training data and a small (200-350) amount of labeled data which we refer to as the \textit{development set}, which we use for (a) a traditional supervision baseline, and (b) for hyperparameter tuning of the end model (see Appendix\versionswitch{}{ \ref{appendix:exp_details}}). 
The average number of weak supervision sources per task was $\AvgNumLFsPerTask$, with sources expressed as Python functions, averaging $\AvgLinesCodePerLF$ lines of code and comprising a mix of pattern matching heuristics, external knowledge base or dictionary lookups, and pre-trained models.
In all three cases, we choose the decomposition into sub-tasks so as to align with weak supervision sources that are either available or natural to express.

\textit{Named Entity Recognition (NER):}
We represent a fine-grained named entity recognition problem---tagging entity mentions in text documents---as a hierarchy of three sub-tasks over the OntoNotes dataset \cite{weischedel2011ontonotes}: $Y_1$ $\in$ $\{\text{Person}, \text{Organization}\}$, $Y_2$ $\in$ $\{\text{Businessperson}, \text{Other Person}, \text{\textit{N/A}}\}$, $Y_3$ $\in$ $\{\text{Company}, \text{Other Org}, \text{\textit{N/A}}\}$, where again we use $\text{\textit{N/A}}$ to represent ``not applicable''.

\textit{Relation Extraction (RE):}
We represent a relation extraction problem---classifying entity-entity relation mentions in text documents---as a hierarchy of six sub-tasks which either concern labeling the subject, object, or subject-object pair of a possible or \textit{candidate} relation in the TACRED dataset~\cite{zhang2017position}.
For example, we might label a relation as having a $\text{Person}$ subject, $\text{Location}$ object, and $\text{Place-of-Residence}$ relation type.

\textit{Medical Document Classification (Doc):}
We represent a radiology report triaging (i.e. document classification) problem from the OpenI dataset~\cite{NationalInstitutesofHealth2017Open-i:Engine} as a hierarchy of three sub-tasks: $Y_1$ $\in$ $\{\text{Acute}, \text{Non-Acute}\}$, $Y_2$ $\in$ $\{\text{Urgent}, \text{Emergent}, \text{\textit{N/A}}\}$, $Y_3$ $\in$ $\{\text{Normal}, \text{Non-Urgent}, \text{\textit{N/A}}\}$.

\paragraph*{End Model Protocol}
Our goal was to test the performance of a basic multi-task end model using training labels produced by various different approaches.
We use an architecture consisting of a shared bidirectional LSTM input layer
with pre-trained embeddings, shared linear intermediate layers, and a separate final linear layer (``task head'') for each task.
Hyperparameters were selected with an initial search for each application (see Appendix), then fixed.

\paragraph*{Core Validation}
We compare the accuracy of the end multi-task model trained with labels from our approach versus those from three baseline approaches (Table~\ref{tab:results}):
\begin{itemize}
  \item \textit{Traditional Supervision} \textbf{[Gold (Dev)]}:
  We train the end model using the small hand-labeled development set.
  
  \item \textit{Hierarchical Majority Vote} \textbf{[MV]}:
  We use a hierarchical majority vote of the weak supervision source labels: i.e. for each data point, for each task we take the majority vote and proceed down the task tree accordingly.
  This procedure can be thought of as a hard decision tree, or a cascade of if-then statements as in a rule-based approach.

  \item \textit{Data Programming} \textbf{[DP]}:
  We model each task separately using the data programming approach for denoising weak supervision \cite{ratner2018snorkel}.
\end{itemize}
In all settings, we used the same end model architecture as described above.
Note that while we choose to model these problems as consisting of multiple sub-tasks, we evaluate with respect to the broad primary task of fine-grained classification (for subtask-specific scores, see Appendix).
We observe in Table~\ref{tab:results} that our approach of leveraging multi-granularity weak supervision leads to large gains---$\AvgGainOverGold$ points over traditional supervision with the development set, $\AvgGainOverMV$ points over hierarchical majority vote, and $\AvgGainOverDP$ points over data programming.

\paragraph*{Ablations}
We examine individual factors:

\textit{Unipolar Correction:}
Modeling unipolar sources (Sec~\ref{sec:extensions}), which we find to be especially common when fine-grained tasks are involved, leads to an average gain of $\AvgUnipolarBoost$ points of accuracy in \systemx performance.

\textit{Joint Task Modeling:}
Next, we use our algorithm to estimate the accuracies of sources for each task separately, to observe the empirical impact of modeling the multi-task setting jointly as proposed.
We see average gains of 1.3 points in accuracy (see Appendix).

\begin{figure}
  \centering
  \includegraphics[width=\figwidthB]{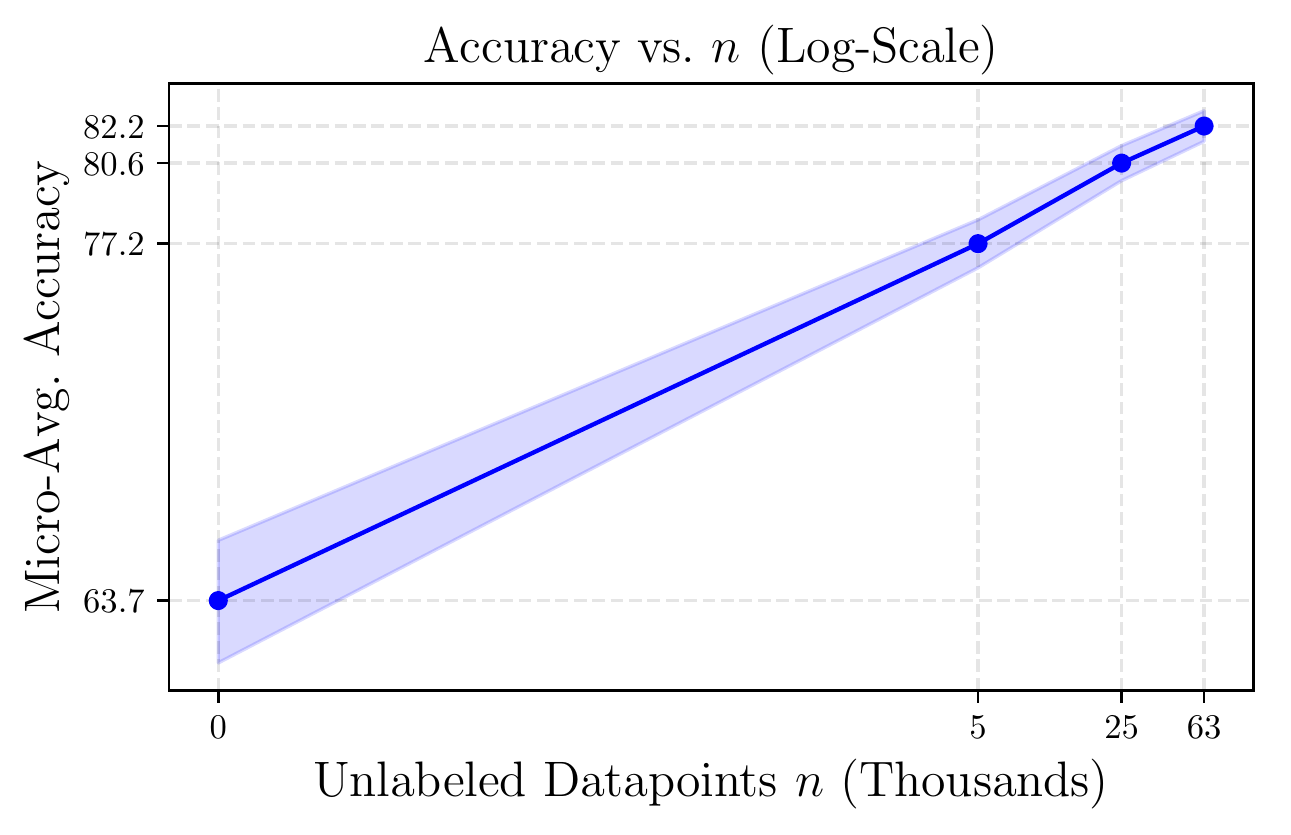}%
  \caption{In the OntoNotes dataset, end model accuracy scales with the amount of available \textit{unlabeled} data.}
  \label{fig:onto_scale}
\end{figure}

\textit{End Model Generalization:}
Though not possible in many settings, in our experiments we can directly apply the label model to make predictions.
In Table~\ref{tab:end_model_boost}, we show that the end model improves performance by an average $\AvgEMOverLM$ points in accuracy, validating that the models trained do indeed learn to generalize beyond the provided weak supervision.
Moreover, the largest generalization gain of 7 points in accuracy came from the dataset with the most available unlabeled data ($n$=63k), demonstrating scaling consistent with the predictions of our theory (Fig.~\ref{fig:onto_scale}).
This ability to leverage additional unlabeled data and more sophisticated end models are key advantages of the weak supervision approach in practice.

\begin{figure}
  \centering
\begin{tabular}{lrrrr}
      \toprule
        & \# Train & LM & EM & \textit{Gain} \\
      \midrule
      NER
        & 62,547
        & 75.2
        & 82.2
        & \textit{7.0}
        \\
      RE
        & 9,090
        & 55.3
        & 57.4
        & \textit{2.1}
        \\
      Doc
        & 2,630
        & 75.6
        & 76.6
        & \textit{1.0}
        \\
      \bottomrule\\
    \end{tabular}
  \caption{Using the label model (LM) predictions directly versus using an end model trained on them (EM).}
  \label{tab:end_model_boost}
\end{figure}

\section{Conclusion}
\label{sec:conclusion}

We presented \systemx, a framework for training models with weak supervision from diverse, \textit{multi-task} sources having different granularities, accuracies, and correlations.
We tackle the core challenge of recovering the unknown source accuracies via a scalable matrix completion-style algorithm, introduce theoretical bounds characterizing the key scaling with respect to unlabeled data, and demonstrate empirical gains on real-world datasets.
In future work, we hope to learn the task relationship structure and cover a broader range of settings where labeled training data is a bottleneck.

\paragraph*{Acknowledgements}
We gratefully acknowledge the support of DARPA under Nos. FA87501720095 (D3M) and FA86501827865 (SDH), NIH under No. N000141712266 (Mobilize), NSF under Nos. CCF1763315 (Beyond Sparsity), CCF1563078 (Volume to Velocity), and DGE-114747 (NSF GRF), ONR under No. N000141712266 (Unifying Weak Supervision), the Moore Foundation, NXP, Xilinx, LETI-CEA, Intel, Google, NEC, Toshiba, TSMC, ARM, Hitachi, BASF, Accenture, Ericsson, Qualcomm, Analog Devices, the Okawa Foundation, and American Family Insurance, the Stanford Interdisciplinary Graduate and Bio-X fellowships, the Intelligence Community Postdoctoral Fellowship, and members of the Stanford DAWN project: Intel, Microsoft, Teradata, Facebook, Google, Ant Financial, NEC, SAP, and VMWare.
The U.S. Government is authorized to reproduce and distribute reprints for Governmental purposes notwithstanding any copyright notation thereon.
Any opinions, findings, and conclusions or recommendations expressed in this material are those of the authors and do not necessarily reflect the views, policies, or endorsements, either expressed or implied, of DARPA, NIH, ONR, or the U.S. Government.

\bibliography{metal}
\bibliographystyle{abbrv}

\pagebreak
\begin{appendix}
  \section{Problem Setup \& Modeling Approach}
  \label{appendix:setup-and-model}
\allowdisplaybreaks

In Section~\ref{appendix:setup-and-model}, we review our problem setup and modeling approach in more detail, and for more general settings than in the body.
In Section~\ref{appendix:theoretical}, we provide an overview, additional interpretation, and the proofs of our main theoretical results.
Finally, in Section~\ref{appendix:exp_details}, we go over additional details of our experimental setup.

We begin in Section~\ref{appendix:glossary} with a glossary of the symbols and notation used throughout this paper.
Then, in Section~\ref{appendix:problem-setup} we present the setup of our multi-task weak supervision problem, and in Section~\ref{appendix:modeling-approach-overview} we present our approach for modeling multi-task weak supervision, and the matrix completion-style algorithm used to estimate the model parameters.
Finally, in Section~\ref{appendix:hierarchical}, we present in more detail the subcase of hierarchical tasks considered in the main body of the paper.

\subsection{Glossary of Symbols}
\label{appendix:glossary}

\begin{table*}[h]
\centering
\begin{tabular}{l l}
\toprule
Symbol & Used for \\
\midrule
$\x$ & Data point, $\x \in \mathcal{X}$ \\
$n$ & Number of data points \\
$Y_s$ & Label for one of the $t$ classification tasks, $Y_s \in \{1, \ldots, k_s\}$ \\
$t$ & Number of tasks \\
$\y$ & Vector of task labels $\textbf{Y} = [Y_1, Y_2, \ldots, Y_t]^T$ \\
$r$ & Cardinality of the output space, $r = |\mathcal{Y}|$ \\
$G_{\text{task}}$ & Task structure graph \\
$\mathcal{Y}$ & Output space of allowable task labels defined by $G_{\text{task}}$, $\y \in \mathcal{Y}$ \\
$\mathcal{D}$ & Distribution from which we assume $(\x, \y)$ data points are sampled i.i.d. \\
$s_i$ & Weak supervision source, a function mapping $\x$ to a label vector\\
$\lf_i$ & Label vector $\lf_i \in \mathcal{Y}$ output by the $i$th source for $\x$ \\
$m$ & Number of sources \\
$\lf$ & $m \times t$ matrix of labels output by the $m$ sources for $\x$ \\
$\mathcal{Y}_0$ & Source output space, which is $\mathcal{Y}$ augmented to include elements set to zero \\
$\tau_i$ & Coverage set of $\lf_i$- the tasks $s_i$ gives non-zero labels to; for convenience, $\tau_0 = \{1,...,t\}$ \\
$\mathcal{Y}_{\tau_i}$ & The output space for $\lf_i$ given coverage set $\tau_i$ \\
$\mathcal{Y}_{\tau_i}^{\text{min}}$ & The output space $\mathcal{Y}_{\tau_i}$ with all but the first value, for defining a minimal set of statistics \\
$G_{\text{source}}$ & Source dependency graph, $G_{\text{source}} = (V,E)$, $V=\{\y,\lf_1,...,\lf_m\}$ \\
$\mathcal{C}$ & Cliqueset (maximal and non-maximal) of $G_{\text{source}}$ \\
$\tilde{\mathcal{C}}, \mathcal{S}$ & The maximal cliques (nodes) and separator sets of the junction tree over $G_{\text{source}}$ \\
$\psi(C, y_C)$ & The indicator variable for the variables in clique $C \in \mathcal{C}$ taking on values $y_C$, $(y_C)_i \in \mathcal{Y}_{\tau_i}$ \\
$\mu$ & The parameters of our label model we aim to estimate; $\mu = \E{}{ \psi }$ \\
$O$ & The set of observable cliques, i.e. those corresponding to cliques without $\y$ \\
$\Sigma$ & Generalized covariance matrix of $O \cup \mathcal{S}$, $\Sigma \equiv \Cov{}{ \psi(O \cup \mathcal{S}) }$ \\
$K$ & The inverse generalized covariance matrix $K = \Sigma^{-1}$ \\
$d_O, d_\mathcal{S}$ & The dimensions of $O$ and $\mathcal{S}$ respectively \\
$G_{\text{aug}}$ & The augmented source dependencies graph $G_{\text{aug}} = (\psi, E_{aug})$ \\
$\Omega$ & The edge set of the inverse graph of $G_{\text{aug}}$ \\
$P$ & Diagonal matrix of class prior probabilities, $P(\y)$ \\
$P_\mu(\y, \lf)$ & The \textit{label model} parameterized by $\mu$ \\
$\tilde{\y}$ & The probabilistic training label, i.e. $P_\mu(\y|\lf)$ \\
$f_w(\x)$ & The \textit{end model} trained using $(\x, \tilde{\y})$ \\
\end{tabular}
\caption{
	Glossary of variables and symbols used in this paper.
}
\label{table:glossary}
\end{table*}

\subsection{Problem Setup}
\label{appendix:problem-setup}
Let $X \in \mathcal{X}$ be a data point and $\y = [Y_1, Y_2, \ldots, Y_t]^T$ be a vector of \textit{task labels} corresponding to $t$ tasks.
We consider categorical task labels, $Y_i \in \{1, \ldots, k_i\}$ for $i \in \{1, \ldots ,t\}$.
We assume $(X, \y)$ pairs are sampled i.i.d. from distribution $\mathcal{D}$; to keep the notation manageable, we do not place subscripts on the sample tuples.

\paragraph*{Task Structure}
The tasks are related by a \emph{task graph} $G_{\text{task}}$.
Here, we consider schemas expressing logical relationships between tasks, which thus define \textit{feasible sets} of label vectors $\mathcal{Y}$, such that $\y \in \mathcal{Y}$.
We let $r = |\mathcal{Y}|$ be the number of feasible task vectors.
In section~\ref{appendix:hierarchical}, we consider the particular subcase of a \textit{hierarchical} task structure as used in the experiments section of the paper.
\paragraph*{Multi-Task Sources}
We now consider \textit{multi-task} weak supervision sources $s_i \in S$, which represent noisy and potentially incomplete sources of labels, which have unknown accuracies and correlations.
Each source $s_i$ outputs label vectors $\lf_i$, which contain non-zero labels for \textit{some} of the tasks, such that $\lf_i$ is in the feasible set $\mathcal{Y}$ but potentially with some elements set to zero, denoting a null vote or abstention for that task.
Let $\mathcal{Y}_0$ denote this extended set which includes certain task labels set to zero.

We also assume that each source has a fixed \textit{task coverage set} $\tau_i$, such that $(\lf_i)_s \neq 0$ for $s \in \tau_i$, and $(\lf_i)_s = 0$ for $s \notin \tau_i$; let $\mathcal{Y}_{\tau_i} \subseteq \mathcal{Y}_0$ be the range of $\lf_i$ given coverage set $\tau_i$.
For convenience, we let $\tau_0 = \{1, \ldots, t\}$ so that $\mathcal{Y}_{\tau_0} = \mathcal{Y}$.
The intuitive idea of the task coverage set is that some labelers may choose not to label certain tasks; Example~\ref{ex:task_structure} illustrates this notion.
Note that sources can also \textit{abstain} for a data point, meaning they emit no label (which we denote with a symbol $\vec{0}$); we include this in $\mathcal{Y}_{\tau_i}$.
Thus we have $s_i : \mathcal{X} \mapsto \mathcal{Y}_{\tau_i}$, where, again, $\lf_i$ denotes the output of the function $s_i$.

\paragraph*{Problem Statement}
Our overall goal is to use the noisy or \textit{weak}, \textit{multi-task} supervision from the set of $m$ sources, $S = \{ s_1, \ldots, s_m \}$, applied to an unlabeled dataset $\mathcal{X}_U$ consisting of $n$ data points, to supervise an \textit{end model} $f_w : \mathcal{X} \mapsto \mathcal{Y}$.
Since the sources have unknown accuracies, and will generally output noisy and incomplete labels that will overlap and conflict, our intermediate goal is to learn a \textit{label model} $P_\mu : \lf \mapsto [0,1]^{|\mathcal{Y}|}$ which takes as input the source labels and outputs a set of probabilistic label vectors, $\tilde{\y}$, for each $X$, which can then be used to train the end model.
Succinctly, given a user-provided tuple $(\mathcal{X}_U, S, G_{\text{source}}, G_{\text{task}})$, our goal is to recover the parameters $\mu$.

The key technical challenge in this approach then consists of learning the parameters of this label model---corresponding to the conditional accuracies of the sources (and, for technical reasons we shall shortly explain, cliques of correlated sources)---given that \textit{we do not have access to the ground truth labels $\y$}.
We discuss our approach to overcoming this core technical challenge in the subsequent section.

\subsection{Our Approach: Modeling Multi-Task Sources}
\label{appendix:modeling-approach-overview}
Our goal is to estimate the parameters $\mu$ of a \textit{label model} that produces probabilistic training labels given the observed source outputs, $\tilde{\y} = P_\mu(\y|\lf)$, \textit{without access to the ground truth labels $\y$}.
We do this in three steps:
\begin{enumerate} 
	\item We start by defining a graphical model over the weak supervision source outputs and the true (latent) variable $\y$, $(\lf_1,\ldots,\lf_m, \y)$, using the conditional independence structure $G_{\text{source}}$ between the sources.
	\item Next, we analyze the \textit{generalized covariance matrix}  $\Sigma$ (following Loh \& Wainwright~\cite{loh2012structure}), which is defined over binary indicator variables for each value of each clique (or specific subsets of cliques) in $G_{\text{source}}$.
	We consider two specific subsets of the cliques in $G_{\text{source}}$, the observable cliques $O$ and the separator sets $\mathcal{S}$, such that:
	\begin{align*}
		\Sigma
		&=
		\begin{bmatrix}
			\Sigma_O & \Sigma_{O\mathcal{S}} \\
			\Sigma_{O\mathcal{S}}^T & \Sigma_\mathcal{S}
		\end{bmatrix}
		&
		\Sigma^{-1}
		&=
		K
		=
		\begin{bmatrix}
			K_O & K_{O\mathcal{S}} \\
			K_{O\mathcal{S}}^T & K_\mathcal{S}
		\end{bmatrix},
	\end{align*}
	where $\Sigma_O$ is the block of $\Sigma$ that we can observe, and $\Sigma_{O\mathcal{S}}$ is a function of $\mu$, the parameters (corresponding to source and  clique accuracies) we wish to recover.
	We then apply a result by Loh and Wainwright \cite{loh2012structure} to establish the sparsity pattern of $K = \Sigma^{-1}$.
	This allows us to apply the block-matrix inversion lemma to reformulate our problem as solving a matrix completion-style objective.
	\item Finally, we describe how to recover the class balance $P(\y)$; with this and the estimate of $\mu$, we then describe how to compute the probabilistic training labels $\tilde{Y} = P_\mu(\y|\lf)$.
\end{enumerate}

We start by focusing on the setting where $G_{\text{source}}$ has a junction tree with singleton separator sets; we note that a version of $G_{\text{source}}$ where this holds can always be formed by adding edges to the graph.
We then discuss how to handle graphs with non-singleton separator sets, and finally describe different settings where our problem reduces to rank-one matrix completion.
In Section~\ref{appendix:theoretical}, we introduce theoretical results for the resulting model and provide our model estimation strategy.

\subsubsection{Defining a Multi-Task Source Model}
\label{appendix:model}

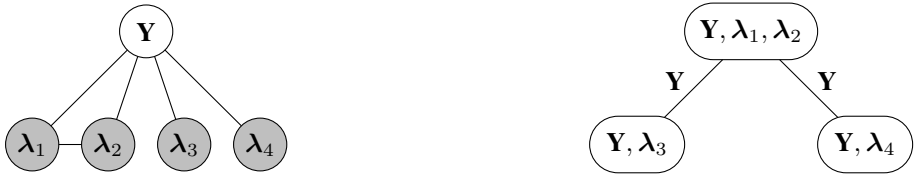
\begin{figure}
	\begin{subfigure}{.5\textwidth}
		\centering
		\begin{tikzpicture}[every node/.style={inner sep=0,outer sep=0}]
		
		\draw (0,1.5) node[draw,circle,fill=white,minimum size=0.7cm](y) {$\y$};
		
		\draw (-1.5,0) node[draw,fill=lightgray,circle,minimum size=0.7cm](l1x) {$\lf_1$};
		\draw (-0.5,0) node[draw,fill=lightgray,circle,minimum size=0.7cm](l2x) {$\lf_2$};
		\draw (0.5,0) node[draw,fill=lightgray,circle,minimum size=0.7cm](l3x) {$\lf_3$};
		\draw (1.5,0) node[draw,fill=lightgray,circle,minimum size=0.7cm](l4x) {$\lf_4$};

		\draw (y) -- (l1x);
		\draw (y) -- (l2x);
		\draw (l1x) -- (l2x);
		\draw (y) -- (l3x);
		\draw (y) -- (l4x);
		
		\end{tikzpicture}
		\label{fig:g-source-appendix}
	\end{subfigure}%
	\begin{subfigure}{.5\textwidth}
		\centering
		\begin{tikzpicture}[every node/.style={inner sep=0,outer sep=0}]
		
		\draw (0,1.5) node [
			draw,
			rounded rectangle,
			fill=white,
			minimum height=0.75cm,
			minimum width=1.75cm
		] (jtn1) {$\y, \lf_1, \lf_2$};
		\draw (-1.5,0) node [
			draw,
			rounded rectangle,
			fill=white,
			minimum height=0.75cm,
			minimum width=1.25cm
		] (jtn2) {$\y, \lf_3$};
		\draw (1.5,0) node [
			draw,
			rounded rectangle,
			fill=white,
			minimum height=0.75cm,
			minimum width=1.25cm
		] (jtn3) {$\y, \lf_4$};

		\draw (jtn1) -- (jtn2);
		\draw (jtn1) -- (jtn3);

		\draw (1,0.85) node [] (ss12) {$\y$};
		\draw (-1,0.85) node [] (ss13) {$\y$};
		
		\end{tikzpicture}
		\label{fig:g-source-jt-appendix}
	\end{subfigure}%
	\caption{
		A simple example of a weak supervision source dependency graph $G_{\text{source}}$ (left) and its junction tree representation (right).
		Here $\y$ is as a vector-valued variable with a feasible set of values, $\y \in |\mathcal{Y}|$, and the output of sources 1 and 2 are modeled as dependent conditioned on $\y$.
		This results in a junction tree with singleton separator sets $\y$.
		Here, the observable cliques are $O = \{\lf_1,\lf_2,\lf_3,\lf_4,\{\lf_1,\lf_2\}\} \subset \mathcal{C}$.
	}
	\label{fig:g-source-example-appendix}
\end{figure}

We consider a model $G_{\text{source}} = (V, E)$, where $V = \{ \y, \lf_1,...,\lf_m \}$, and $E$ consists of pairwise interactions (i.e. we consider an Ising model, or equivalently, a graph rather than a hypergraph of correlations).
We assume that $G_{\text{source}}$ is provided by the user.
However, if $G_{\text{source}}$ is unknown, there are various techniques for estimating it statistically~\cite{bach2017learning} or even from static analysis if the sources are heuristic functions~\cite{varma2017inferring}.
We provide an example $G_{\text{source}}$ with singleton separator sets in Figure~\ref{fig:g-source-example-appendix}.

\paragraph*{Augmented Sufficient Statistics}
Finally, we extend the random variables in $V$ by defining a matrix of indicator statistics over all cliques in $G_{\text{source}}$, in order to estimate all the parameters needed for our label model $P_\mu$.
We assume that the provided $G_{\text{source}}$ is \textit{chordal}, meaning it has no chordless cycles of length greater than three; if not, the graph can easily be \emph{triangulated} to satisfy this property, in which case we work with this augmented version.

Let $\mathcal{C}$ be the set of maximal and non-maximal cliques in the chordal graph $G_{\text{source}}$.
We start by defining a binary indicator random variable for the event of a clique $C \in \mathcal{C}$ in the graph $G_{\text{source}} = (V, E)$ taking on a set of values $y_C$:
\begin{align*}
	\psi(C,y_C)
	&=
	\ind{ \cap_{i \in C} V_i = (y_C)_i },
\end{align*}
where $(y_C)_i \in \mathcal{Y}_{\tau_i}^{\text{min}}$ and $\mathcal{Y}_{\tau_i}^{\text{min}}$ contains all but one values of $\mathcal{Y}_{\tau_i}$, thereby leading to a minimal set of statistics.
Note that in our notation, $V_0 = \y$, $\mathcal{Y}_{\tau_0} = \mathcal{Y}$, and $V_{i > 0} = \lf_i$.
Accordingly, we define $\psi(C) \in \{0,1\}^{\prod_{i\in C} (|\mathcal{Y}_{\tau_i}|-1)}$ as the vector of indicator random variables for all combinations of all but one of the labels emitted by each variable in clique $C$, and define $\psi(\textbf{C})$ accordingly for any set of cliques $\textbf{C} \subseteq \mathcal{C}$.
Then $\mu = \E{}{ \psi(\mathcal{C}) }$ is the vector of sufficient statistics for the label model we want to learn.
Our model estimation goal is now stated simply: we wish to estimate $\mu$, \textit{without access to the ground truth labels} $\y$.

\subsubsection{Model Estimation without Ground Truth Using Inverse Covariance Structure}
\label{appendix:model-estimation}
Our goal is to estimate $\mu = \E{}{ \psi(\mathcal{C}) }$; this, along with the class balance $P(\y)$ (which we assume we know, or else estimate using the approach in Section~\ref{appendix:recovering-class-balance}), is sufficient information to compute $P_\mu(\y|\lf)$.
If we had access to a large enough set of ground truth labels $\y$, we could simply take the empirical expectation $\Ehat{ \psi }$; however in our setting we cannot directly observe this.
Instead, we proceed by analyzing a sub-block of the covariance matrix of $\psi(\mathcal{C})$, which corresponds to the \textit{generalized covariance matrix} of our graphical model as in~\cite{loh2012structure}, and leverage two key pieces of information:
\begin{itemize}
	\item A sub-block of this generalized covariance matrix is observable, and
	\item By a simple extension of Corollary 1 in~\cite{loh2012structure}, we know the sparsity structure of the inverse generalized covariance matrix $\Sigma^{-1}$, i.e. we know that it will have elements equal to zero according to the structure of $G_{\text{source}}$.
\end{itemize}
Since $G_{\text{source}}$ is triangulated, it admits a \textit{junction tree} representation~\cite{koller2009probabilistic}, which has maximal cliques (nodes) $\tilde{\mathcal{C}}$ and separator sets $\mathcal{S}$.
Note that we follow the convention that $\mathcal{S}$ includes the full powerset of separator set cliques, i.e. all subset cliques of separator set cliques are also included in $\mathcal{S}$.
We proceed by considering two specific subsets of the cliques of our graphical model $G_{\text{source}}$: those that are observable (i.e. not containing $\y$), $O = \{ C ~|~ \y \notin C, C \in \mathcal{C} \}$, and the set of separator set cliques (which will always contain $\y$, and thus be unobservable).

For simplicity of exposition, we start by considering graphs $G_{\text{source}}$ which have singleton separator sets; given our graph structure, this means that $\mathcal{S} = \{\{\y\}\}$.
Note that in general we will write single-element sets without braces when their type is obvious from context, so we have $\mathcal{S} = \{\y\}$.
Intuitively, this corresponds to models where weak supervision sources are correlated in fully-connected clusters, corresponding to real-world settings in which sources are correlated due to shared data sources, code, or heuristics.
However, we can always either (i) add edges to $G_{\text{source}}$ such that this is the case, or (ii) extend our approach to many settings where $G_{\text{source}}$ does not have singleton separator sets (see Section~\ref{appendix:non-singleton-sep-sets}).

In this singleton separator set setting of $\mathcal{S} = \{\y\}$, we now have:
\begin{align*}
	O
	&=
	\{ C ~|~ \y \notin C, C \in \mathcal{C} \}
	&
	\mathcal{S} = \{ \y \}.
\end{align*}
where $\psi(O)$ and $\psi(\y)$ are the corresponding vectors of minimal indicator variables.
We define corresponding dimensions $d_O$ and $d_\mathcal{S}$:
\begin{align*}
	d_O
	&=
	\sum_{C \in O} \prod_{i \in C} (|\mathcal{Y}_{\tau_i}|-1)
	&
	d_\mathcal{S}
	&=
	r-1.
\end{align*}

We now decompose the generalized covariance matrix and its inverse as:
\begin{align}
	\label{eqn:gen-cov}
	\Cov{}{ \psi(O \cup \mathcal{S}) }
	&\equiv
	\Sigma
	=
	\begin{bmatrix}
		\Sigma_O & \Sigma_{O\mathcal{S}} \\
		\Sigma_{O\mathcal{S}}^T & \Sigma_\mathcal{S}
	\end{bmatrix}
	&
	\Sigma^{-1}
	&=
	K
	=
	\begin{bmatrix}
		K_O & K_{O\mathcal{S}} \\
		K_{O\mathcal{S}}^T & K_\mathcal{S}
	\end{bmatrix},
\end{align}
This is similar to the form used in~\cite{chandrasekaran2010latent}, but with several important differences: we consider discrete (rather than Gaussian) random variables and have additional knowledge of the graph structure.
Here, $\Sigma_O$ is the observable block of the generalized covariance matrix $\Sigma$, and $\Sigma_{O\mathcal{S}}$ is the unobserved block which is a function of $\mu$, the parameters (corresponding to source and source clique accuracies) we wish to recover.
Note that with the singleton separator sets we are considering, $\Sigma_\mathcal{S}$ is a function of the class balance $P(\y)$, which we assume is either known, or has been estimated according to the unsupervised approach we detail in Section~\ref{appendix:recovering-class-balance}.
Therefore, we assume that $\Sigma_\mathcal{S}$ is also known.
Concretely then, our goal is to recover $\Sigma_{O\mathcal{S}}$ given $\Sigma_O, \Sigma_\mathcal{S}$.

We start by applying the block matrix inversion lemma to get the equation:
\begin{align}
	\label{eqn:block-inv-cov}
	K_O
	&=
	\Sigma_O^{-1}
	+ \Sigma_O^{-1}\Sigma_{O\mathcal{S}}\left( \Sigma_\mathcal{S} - \Sigma_{O\mathcal{S}}^T\Sigma_O^{-1}\Sigma_{O\mathcal{S}} \right)^{-1} \Sigma_{O\mathcal{S}}^T\Sigma_O^{-1}.
\end{align}

Next, let $JJ^T = \left( \Sigma_\mathcal{S} - \Sigma_{O\mathcal{S}}^T\Sigma_O^{-1}\Sigma_{O\mathcal{S}} \right)^{-1}$.
We justify this decomposition by showing that this term is positive semidefinite.
We start by applying the Woodbury matrix inversion lemma:
\begin{align}
	\label{eqn:middle-term}
	\left( \Sigma_\mathcal{S} - \Sigma_{O\mathcal{S}}^T\Sigma_O^{-1}\Sigma_{O\mathcal{S}} \right)^{-1}
	&=
	\Sigma_\mathcal{S}^{-1}
	+ \Sigma_\mathcal{S}^{-1}\Sigma_{O\mathcal{S}}^T \left( \Sigma_O + \Sigma_{O\mathcal{S}}\Sigma_\mathcal{S}^{-1}\Sigma_{O\mathcal{S}}^T \right)^{-1} \Sigma_{O\mathcal{S}}\Sigma_\mathcal{S}^{-1}.
\end{align}
Now, note that $\Sigma_O$ and $\Sigma_\mathcal{S}$ are both covariance matrices themselves and are therefore PSD.
Furthermore, from~\cite{loh2012structure} we know that $\Sigma^{-1}$ must exist, which implies that $\Sigma_O$ and $\Sigma_\mathcal{S}$ are invertible (and thus in fact positive definite).
Therefore we also have that $\Sigma_{O\mathcal{S}}\Sigma_\mathcal{S}^{-1}\Sigma_{O\mathcal{S}}^T \succ 0 \implies \left( \Sigma_O + \Sigma_{O\mathcal{S}}\Sigma_\mathcal{S}^{-1}\Sigma_{O\mathcal{S}}^T \right)^{-1} \succ 0$, and therefore (\ref{eqn:middle-term}) is positive definite, and can therefore always be expressed as $JJ^T$ for some $J$.
Therefore, we can write (\ref{eqn:block-inv-cov}) as:
\begin{align*}
	K_O
	&=
	\Sigma_O^{-1}
	+ \Sigma_O^{-1}\Sigma_{O\mathcal{S}} JJ^T \Sigma_{O\mathcal{S}}^T\Sigma_O^{-1}.
\end{align*}
Finally, define $Z = \Sigma_O^{-1}\Sigma_{O\mathcal{S}} J$; we then have:
\begin{align}
	\label{eqn:matrix-completion-form}
	K_O
	&=
	\Sigma_O^{-1} + ZZ^T.
\end{align}
Note that $Z \in \mathbb{R}^{d_O \times d_H}$, where $d_H = r-1$, and therefore $ZZ^T$ is a rank-$(r-1)$ matrix.
Therefore, we now have a form (\ref{eqn:matrix-completion-form}) that appears close to being a matrix completion-style problem.
We complete the connection by leveraging the known sparsity structure of $K_O$.

Define $G_{\text{aug}} = (\psi, E_{\text{aug}})$ to be the augmented version of our graph $G_{\text{source}}$.
In other words, let $i = (C_1, y_{C_1})$ and $j = (C_2, y_{C_2})$ according to the indexing scheme of our augmented indicator variables; then, $(i,j) \in E_{\text{aug}}$ if $C_1, C_2$ are subsets of the same maximal clique in $G_{\text{source}}$.
Then, let $G_{\text{inv-aug}} = (\psi, \Omega)$ be the inverse graph of $G_{\text{aug}}$, such that $(i,j) \in  E_{\text{aug}} \implies (i,j) \notin \Omega$ and vice-versa.

We start with a result that extends Corollary 1 in Loh \& Wainwright~\cite{loh2012structure} to our specific setting where we consider a set of the variables that contains all observable cliques, $O$, and all separator sets $\mathcal{S}$ (note that this result holds for all $\mathcal{S}$, not just $\mathcal{S} = \{\y\}$):

\begin{corollary}
	\label{cor:o-h-partition}
	Let $U = O \cup \mathcal{S}$.
	Let $\Sigma_U$ be the generalized covariance matrix for $U$.
	Then $(\Sigma_U^{-1})_{i,j} = 0$ whenever $i,j$ correspond to cliques $C_1, C_2$ respectively such that $C_1, C_2$ are not subsets of the same maximal clique.
\end{corollary}
\textit{Proof:}
We partition the cliques $\mathcal{C}$ into two sets, $U$ and $W = \mathcal{C} \setminus U$.
Let $\Sigma$ be the full generalized covariance matrix (i.e. including all maximal and non-maximal cliques) and $\Gamma = \Sigma^{-1}$.
Thus we have:
\begin{align*}
	\Sigma
	&=
	\begin{bmatrix}
		\Sigma_U & \Sigma_{UW} \\
		\Sigma_{UW}^T & \Sigma_W
	\end{bmatrix}
	&
	\Sigma^{-1}
	=
	\Gamma
	=
	\begin{bmatrix}
		K_U & K_{UW} \\
		K_{UW}^T & K_W
	\end{bmatrix}.
\end{align*}
By the block matrix inversion lemma we have:
\begin{align*}
	\Sigma_U^{-1}
	&=
	K_U - K_{UW} K_W^{-1} K_{UW}^T.
\end{align*}
We now follow the proof structure of Corollary 1 of~\cite{loh2012structure}.
We know $K_U$ is graph structured by Theorem 1 of~\cite{loh2012structure}.
Next, using the same argument as in the proof of Corollary 1 of~\cite{loh2012structure}, we know that $K_W$, and therefore $K_W^{-1}$, is block-diagonal.
Intuitively, because the set $U$ contains all of the separator set cliques, and due to the running intersection property of a junction tree, each clique in $W$ belongs to precisely one maximal clique- leading to block diagonal structure of $K_W$.
We thus need only to show that the following quantity is zero for two cliques $C_i, C_j$ that are not subsets of the same maximal clique, with corresponding indices $i,j$:
\begin{align*}
	\left( K_{UW} K_W^{-1} K_{UW}^T \right)_{i,j}
	&=
	\sum_{B} (K_{UW})_{i,B} (K_W^{-1})_{B,B} (K_{UW}^T)_{B,j},
\end{align*}
where $B$ are the indices corresponding to the blocks in $K_W^{-1}$, which correspond to maximal cliques.
Our argument follows again as in Corollary 1 of~\cite{loh2012structure}: since $U$ contains the separator sets, if the two cliques $C_1, C_2$ are not subsets of the same maximal clique, then for each $B$, either $(K_{UW})_{i,B}$ or $(K_{UW}^T)_{B,j}$ must be zero, completing the proof.

Now, by Corollary 1, we know that $K_{i,j} = 0$ if $(i,j) \in \Omega$.
Let $A_\Omega$ denote a matrix $A$ with all entries $(i,j) \notin \Omega$ masked to zero.
Then, we have:
\begin{align}
	\left( \Sigma_O^{-1} \right)_\Omega + \left( ZZ^T \right)_\Omega &= 0.
\end{align}

Thus, given the dependency graph $G_{\text{source}}$, we can solve for $Z$ as a rank-$(r-1)$ matrix completion problem, with mask $\Omega$.
Defining the semi-norm $\norm{ A }_\Omega = \norm{ A_\Omega }_F$, we can solve:
\begin{align}
	\hat{Z}
	&=
	\argmin{Z}{ \norm{ \Sigma_O^{-1} + Z Z^T }_\Omega }.
\end{align}

Now, we have an estimate of $Z$.
Note that at this point, we can only recover $Z$ up to orthogonal transformations.
We proceed by considering a reduced rank-one model, detailed in Section~\ref{appendix:rank-one-reduction}, and in Section~\ref{appendix:identifiability} establish concrete conditions under which this model is uniquely identifiable.

We denote this rank-one setting by switching to writing $Z$ as $z \in \mathbb{R}^{d_O \times 1}$, in which case we now have:
\begin{align}
	\hat{z}
	&=
	\argmin{z}{ \norm{ \Sigma_O^{-1} + zz^T }_\Omega }.
\end{align}
Once we have recovered $z$ uniquely (see Section~\ref{appendix:identifiability}), we next need to recover $\Sigma_{O\mathcal{S}} = c^{-\frac12}\Sigma_O z$.
We use the fact that $c = \Sigma_{\mathcal{S}}^{-1}(1 + z^T\Sigma_Oz)$, which we can confirm explicitly below, starting from the definition of $c$:
\begin{align*}
	c
	&=
	\left( 
		\Sigma_{\mathcal{S}}
		- \Sigma_{O\mathcal{S}}^T\Sigma_O^{-1}\Sigma_{O\mathcal{S}}
	\right)^{-1}\\
	&=
	\left( 
		\Sigma_{\mathcal{S}}
		- (c^{-\frac12}\Sigma_O z)^T\Sigma_O^{-1}(c^{-\frac12}\Sigma_O z)
	\right)^{-1}\\
	&=
	\left( 
		\Sigma_{\mathcal{S}}
		- c^{-1} z^T\Sigma_Oz
	\right)^{-1}\\
	\implies
	c^{-1}
	&=
	\Sigma_{\mathcal{S}} - c^{-1} z^T\Sigma_Oz\\
	\implies
	c^{-1}\left(1 + z^T\Sigma_Oz\right)
	&=
	\Sigma_{\mathcal{S}}\\
	\implies
	c
	&=
	\Sigma_{\mathcal{S}}^{-1}\left(1 + z^T\Sigma_Oz\right)
\end{align*}
Thus, we can directly recover an estimate of $\Sigma_{O\mathcal{S}}$ from the observed $\Sigma_O$, known $\Sigma_{\mathcal{S}}$, and estimated $z$.
Finally, we have:
\begin{align}
	\Sigma_{O\mathcal{S}} + \E{}{ \psi(O) }\E{}{ \psi(\mathcal{S}) }^T
	&=
	\E{}{ \psi(O)\psi(\mathcal{S})^T }.
\end{align}
Here, we can clearly observe $\E{}{ \psi(O) }$, and given that we know the class balance $P(\y)$, we also have $\E{}{ \psi(\mathcal{S}) }$; therefore we can compute $\E{}{ \psi(O) \psi(\mathcal{S})^T }$.
Our goal now is to recover the columns $\E{}{ \psi(O)\psi(\y_i) }$, which together make up $\mu$; we can do this based on the constraints of our rank-one model (Section~\ref{appendix:rank-one-reduction}), thus recovering an estimate of $\mu$, which given the uniqueness of $\hat{z}$ (Section~\ref{appendix:identifiability}) is also unique.
The overall procedure is described in the main body, in Algorithm~\ref{alg:method}.

\subsubsection{Handling Non-Singleton Separator Sets}
\label{appendix:non-singleton-sep-sets}

Now, we consider the setting where $G_{\text{source}}$ has arbitrary separator sets.
Let $d_S = \sum_{S \in \mathcal{S}} \prod_{i \in S} (|\mathcal{Y}_{\tau_i}|-1)$.
We see that we could solve this using our standard approach---this time, involving a rank-$d_S$ matrix completion problem---except for the fact that we do not know $\Sigma_\mathcal{S}$, as it now involves terms besides the class balance.

Note first of all that we can always add edges between sources to $G_{\text{source}}$ such that it has singleton separator sets (intuitively, this consists of ``completing the clusters''), and as long as our problem is still identifiable (see Section~\ref{appendix:identifiability}), we can simply solve this instance as above.

Instead, we can also take a multi-step approach, wherein we first consider one or more subgraphs of $G_{\text{source}}$ that contain only singleton separator sets, and contain the cliques in $\mathcal{S}$.
We can then solve this problem as before, which then gives us the needed information to identify the elements of $\Sigma_\mathcal{S}$ in our full problem, which we can then solve.
In particular, we see that this multi-step approach is possible whenever the graph $G_{\text{source}}$ has at least three components that are disconnected except for through $\y$.

\subsubsection{Rank-One Settings}
\label{appendix:rank-one-reduction}

We now consider settings where we can estimate the parameters of our label model, $\mu$, involving only a rank-one matrix completion problem.

First, in the simplest setting of a single-task problem with binary class variable, $\y \in \{0,1\}$ and $G_{\text{source}}$ with singleton separator sets, $d_H = r-1 = 1$ and our problem is directly a rank-one instance.

Next, we consider the setting of general $\y$, with $|\mathcal{Y}| = r$ and $G_{\text{source}}$ with singleton separator sets.
By default, our problem now involves a rank-$(r-1)$ matrix completion problem.
However, we can reduce this to involving only a rank-one matrix completion problem by adding one simplifying assuption to our model: namely, that sources emit different incorrect labels with uniform conditional probability.
Concretely, we add the assumption that:
\begin{align}
	\label{eqn:simplified-model-appendix}
	(\lf_C)_i = \y \iff (\lf'_C)_i = \y ~~~ \forall i \in C
	& \implies
	P(\lf_C | \y)
	=
	P(\lf'_C | \y)
\end{align}
Note that this is the same assumption as in the main body, but expressed more explicitly with respect to a clique $C$.
For example, under this assumption, $P(\lf_i=y'|\y=y)$ is the same for all $y'$ such that $y' \neq y$.
As another example, $P(\lf_i=y, \lf_j=y'|\y=y)$ is the same for all $y'$ such that $y' \neq y$.
Intuitively, under this commonly-used model, we are not modeling the different class-wise errors a source makes, but rather just whether it is correct or not given the correctness of other sources it is correlated with.
The idea then is that with assumption (\ref{eqn:simplified-model-appendix}) even though $|H| = r - 1$ (and thus $\Sigma_{O\mathcal{S}}$ has $r-1$ columns), we only actually need to solve for a single parameter per element of $O$.

We can operationalize this by forming a new graph with a binarized version of $\y$, $\y_B \in \{0,1\}$, such that the $r$ classes are mapped to either $0$ or $1$.
We see that this new variable still results in the same structure of dependencies $G_{\text{source}}$, and still allows us to recover the parameters $\alpha_y$ (and thus $\mu$).
We now have:
\begin{align*}
	\mathcal{S}
	&=
	\{ \y_B \}
\end{align*}
We now solve in the same rank-one way as in the binary $\y$ case.
Now, for singleton cliques, $\{ \lf_i, \y \}$, given that we know $P(\y)$, we can directly recover $P(\lf_i=y | \y=y')$ for all $y'$, given our simplified model.

For non-singleton cliques $\{ \lf_C, \y \}$, note that we can directly recover $P(\cap_{i \in C} \lf_i=y|\y = y')$ in the exact same way.
From these, computed for all cliques, we can then recover any probability in our model.
For example, for $y' \neq y$:
\begin{align*}
	P(\lf_i=y, \lf_j=y' | \y=y)
	&=
	P(\lf_i=y | \y=y) - \sum_{y''\neq y'} P(\lf_i=y, \lf_j=y'' | \y=y)\\
	&=
	P(\lf_i=y | \y=y) - P(\lf_i=y, \lf_j=y | \y=y) - \\
&\qquad \qquad \times (r-2) P(\lf_i=y, \lf_j=y' | \y=y)\\
	\implies
	P(\lf_i=y, \lf_j=y' | \y=y)
	&=
	\frac{1}{r-1} \left(
		P(\lf_i=y | \y=y) - P(\lf_i=y, \lf_j=y | \y=y)
	\right).
\end{align*}
In this way, we can recover all of the parameters $\mu$ while only involving a rank-one matrix completion problem.
Note that this also suggests a way to solve for the more general model, i.e. without (\ref{eqn:simplified-model-appendix}), using a hierarchical classification approach.

\subsubsection{Recovering the Class Balance $P$ \& Computing $P(Y|\lambda)$}
\label{appendix:recovering-class-balance}
We now turn to the task of recovering the class balance $P(\y)$, for $\y \in \mathcal{Y}$.
In many practical settings, $P(\y)$ can be estimated from a small labeled sample, or may be known in advance.
However here, we consider using a subset of conditionally independent sources, $s_1, \ldots, s_k$ to estimate $P(\y)$.
We note first of all that simply taking the majority vote of these sources is a biased estimator.

Instead, we consider a simplified version of the matrix completion-based approach taken so far.
Here, we consider a subset of the sources $s_1,\ldots ,s_k$ such that they are conditionally independent given $G_{\text{source}}$, i.e. $\lf_i \perp \lf_j | \y$, and consider only the unary indicator statistics.
Denote the vector of these unary indicator statistics over the conditionally independent subset of sources as $\phi$, and let the observed overlaps matrix between sources $i$ and $j$ be $A_{i,j} = \E{}{ \phi_i\phi_j^T }$.
Note that due to the conditional independence of $\lf_i$ and $\lf_j$, for any $k,l$ we have:
\begin{align*}
	(A_{i,j})_{k,l}
	&=
	\E{}{ (\phi_i)_k (\phi_j)_l } \\
	&=
	P(\lf_i=y_k, \lf_j=y_l) \\
	&=
	\sum_{y \in \mathcal{Y}} P(\lf_i=y_k, \lf_j=y_l | \y=y) P(\y=y) \\
	&=
	\sum_{y \in \mathcal{Y}} P(\lf_i=y_k | \y=y) P(\lf_j=y_l | \y=y) P(\y=y).
\end{align*}
Letting $B_i$ be the $|\mathcal{Y}_{\tau_i}| \times |\mathcal{Y}|$ matrix of conditional probabilities, $(B_i)_{j,k} = P(\lf_i=y_j | \y=y_k)$, and $P$ be the diagonal matrix such that $P_{i,i} = P(\y=y_i)$, we can re-express the above as:
\begin{align*}
	A_{i,j}
	&=
	B_i P B_j^T.
\end{align*}
Since $P$ is composed of strictly positive elements, and is diagonal (and thus PSD), we re-express this as:
\begin{align}
	\label{eqn:ind-decomp}
	A_{i,j}
	&=
	\tilde{B}_i \tilde{B}_j^T,
\end{align}
where $\tilde{B}_i = B_i \sqrt{P}$.
We could now try to recover $P$ by decomposing the observed $A_{i,j}$ to recover the $\tilde{B}_i$, and from there recover $P$ via the relation:
\begin{align}
	\label{eqn:p-recovery}
	P
	&=
	\text{diag}\left( \tilde{B}_i^T\vec{1} \right)^2,
\end{align}
since summing the column of $\tilde{B}_i$ corresponding to label $\y$ is equal to $\sqrt{P(\y)} \sum_{y \in \mathcal{Y}_i} P(\lf_i = y | \y) = \sqrt{P(\y)}$ by the law of total probability.
However, note that $\tilde{B}_i U$ for any orthogonal matrix $U$ also satisfies~(\ref{eqn:ind-decomp}), and could thus lead to a potentially infinite number of incorrect estimates of $P$.

\paragraph*{Class Balance Identifiability with Three-Way View Constraint}
A different approach involves considering the three-way overlaps observed as $A_{i,j,k}$.
This is equivalent to performing a tensor decomposition.
Note that above, the problem is that matrix decomposition is typically invariant to rotations and reflections; tensor decompositions have easier-to-meet uniqueness conditions (and are thus more rigid).

Specifically, we apply Kruskal's classical identifiability condition for unique 3-tensor decomposition.
Consider some tensor 
\[T = \sum_{r=1}^R X_r \otimes Y_r \otimes Z_r,\]
where $X_r, Y_r, Z_r$ are column vectors that make up the matrices $X, Y, Z$. 
The Kruskal rank $k_X$ of $X$ is the largest $k$ such that any $k$ columns of $X$ are linearly independent.
Then, the decomposition above is unique if $k_X + k_Y + k_Z \geq 2R+2$~\cite{kruskal77, bhaskara14}.
In our case, our triple views have $R=|\mathcal{Y}|$, and we have
\begin{align}
	\label{eqn:ind-tensor-decomp}
	A_{i,j,k}
	&=
	\tilde{B}_i \otimes \tilde{B}_j \otimes \tilde{B}_k.
\end{align}
Thus, if $k_{\tilde{B}_i} + k_{\tilde{B}_j} + k_{\tilde{B}_k} \geq 2|\mathcal{Y}| + 2$, we have identifiability.
Thus, it is sufficient to have the columns of each of the $\tilde{B}_i$'s be linearly independent.
Note that each of the $B_i$'s have columns with the same sum, so these columns are only linearly dependent if they are equal, which would only be the case if the sources were random voters.

Thus, we can use (\ref{eqn:ind-tensor-decomp}) to recover the $\tilde{B}_i$ in a stable fashion, and then use (\ref{eqn:p-recovery}) to recover the $P(\y)$.

\subsubsection{Predicting Labels with the Label Model}
\label{appendix:label-model-predictions}
Once we have an estimate of $\mu$, we can make predictions with the label model---i.e. generate our \textit{probabilistic} training labels $P_\mu(\y|\lf)$---using the junction tree we have already defined over $G_{\text{source}}$.
Specifically, let $\tilde{\mathcal{C}}$ be the set of maximal cliques (nodes) in the junction tree, and let $\mathcal{S}$ be the set of separator sets.
Then we have:
\begin{align*}
	P_\mu(\y, \lf)
	&=
	\frac{
		\prod_{C \in \tilde{\mathcal{C}}} P(V_C)
	}{
		\prod_{S \in \mathcal{S}} P(V_S)
	}
	=
	\frac{
		\prod_{C \in \tilde{\mathcal{C}}} \mu_{(C, (\y,\lf_C))}
	}{
		\prod_{S \in \mathcal{S}} \mu_{(S, (\y,\lf_S))}
	},
\end{align*}
where again, $V_C = \{ V_i \}_{i \in C}$, where $V_0 = \y$ and $V_{i>0} = \lf_i$.
Thus, we can directly compute the predicted labels $P_\mu(\y|\lf)$ based on the estimated parameters $\mu$.

\subsection{Example: Hierarchical Multi-Task Supervision}
\label{appendix:hierarchical}
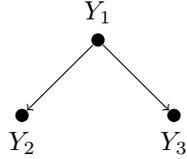
\begin{figure}
\centering
\begin{tikzpicture}[scale=1.0]
\node [circle,fill,inner sep=0pt,minimum size=5pt] (a0) at (0,1) [label={above:$Y_1$}]{};
\node [circle,fill,inner sep=0pt,minimum size=5pt] (a1) at (-1,0) [label={below:$Y_2$}]{};
\node [circle,fill,inner sep=0pt,minimum size=5pt] (a2) at (1,0) [label={below:$Y_3$}]{};
\draw[->] (a0) -- (a1);
\draw[->](a0) -- (a2);
\end{tikzpicture}
\caption{
	Example task hierarchy $G_{\text{task}}$ for a three-task classification problem.
	Task $Y_1$ classifies a data point $X$ as a \texttt{PERSON} or \texttt{BUILDING}.
	If $Y_1$ classifies $X$ as a \texttt{PERSON}, $Y_2$ is used to distinguish between \texttt{DOCTOR} and \texttt{NON-DOCTOR}.
	Similarly, if $Y_2$ classifies $X$ as a \texttt{BUILDING}, $Y_3$ is used to distinguish between \texttt{HOSPITAL} and \texttt{NON-HOSPITAL}.
	Tasks $Y_2,Y_3$ are more specific, or \textit{finer-grained} tasks, constrained by their parent task $Y_1$.
}
\label{fig:tree_task}
\end{figure}

We now consider the specific case of \textit{hierarchical} multi-task supervision, which can be thought of as consisting of coarser- and finer-grained labels, or alternatively higher- and lower-level labels, and provides a way to supervise e.g. fine-grained classification tasks at multiple levels of granularity.
Specifically, consider a task label vector $\y = [Y_1, \ldots, Y_t]^T$ as before, this time with $Y_s \in \{\textit{\texttt{N/A}}, 1, \ldots, k_s\}$, where we will explain the meaning of the special value $\textit{\texttt{N/A}}$ shortly.
We then assume that the tasks $Y_s$ are related by a \textit{task hierarchy} which is a hierarchy $G_{\text{task}} = (V,E)$ with vertex set $V = \{Y_1, Y_2, \ldots, Y_t\}$ and directed edge set $E$.
 The task structure reflects constraints imposed by higher level (more general) tasks on lower level (more specific) tasks.
 The following example illustrates a simple tree  task structure:

\begin{example}
Let $Y_1$ classify a data point $X$ as either a \texttt{PERSON} ($Y_1 = 1$) or \texttt{BUILDING} ($Y_1 = 2$).
If $Y_1 = 1$, indicating that $X$ represents a \texttt{PERSON}, then $Y_2$ can further label $X$ as a \texttt{DOCTOR} or \texttt{NON-DOCTOR}.
$Y_3$ is used to distinguish between \texttt{HOSPITAL} and \texttt{NON-HOSPITAL} in the case that $Y_1 = 2$.
The corresponding graph $G_{\text{task}}$ is shown in Figure~\ref{fig:tree_task}.
If $Y_1=2$, then task $Y_2$ is not applicable, since $Y_2$ is only suitable for persons; in this case, $Y_2$ takes the value $\textit{\texttt{N/A}}$.
In this way the task hierarchy defines a feasible set of task vector values: $ \y = [1, 1, \textit{\texttt{N/A}}]^T, [1, 2, \textit{\texttt{N/A}}]^T, [2,\textit{\texttt{N/A}}, 1]^T, [2,\textit{\texttt{N/A}}, 2]^T$ are valid, while e.g. $\y = [1, 1, 2]^T$ is not.
\label{ex:task_structure}
\end{example}

As in the example, for certain configurations of $\y$'s, the parent tasks logically constrain the one or more of the children tasks to be irrelevant, or rather, to have inapplicable label values.
In this case, the task takes on the value $\textit{\texttt{N/A}}$.
In Example~\ref{ex:task_structure}, we have that if $Y_1 = 1$, representing a building, then $Y_2$ is inactive (since $X$ corresponds to a building).
We define the symbol $\textit{\texttt{N/A}}$ (for incompatible) for this scenario.
More concretely, let $\mathcal{N}(Y_i) = \{Y_j : (Y_j, Y_i) \in E\}$ be the in-neighborhood of $Y_i$.
Then, the values of the members of $\mathcal{N}(Y_i)$ determine whether $Y_i = \textit{\texttt{N/A}}$, i.e., $\mathds{1}\{Y_j = \textit{\texttt{N/A}}\}$ is deterministic conditioned on $\mathcal{N}(Y_i)$. 
\paragraph*{Hierarchical Multi-Task Sources}
Observe that in the mutually-exclusive task hierarchy just described, the value of a descendant task label $Y_d$ determines the values of all other task labels in the hierarchy besides its descendants.
For example, in Example~\ref{ex:task_structure}, a label $Y_2 = 1 \implies (Y_1=1, Y_3=\textit{\texttt{N/A}})$; in other words, knowing that $X$ is a \texttt{DOCTOR} also implies that $X$ is a \texttt{PERSON} and not a \texttt{BUILDING}.

For a source $\lf_i$ with coverage set $\tau_i$, the label it gives to the lowest task in the task hierarchy which is non-zero and non-$\textit{\texttt{N/A}}$ determines the entire label vector output by $\lf_i$.
E.g. if the lowest task that $\lf_i$ labels in the hierarchy is $Y_1=1$, then this implies that it outputs vector $[1, 0, \textit{\texttt{N/A}}]^T$.
Thus, in this sense, we can think of each sources $\lf_i$ as labeling one specific task in the hierarchy, and thus can talk about coarser- and finer-grained sources.
\paragraph*{Reduced-Rank Form: Modeling Local Accuracies}
In some cases, we can make slightly different modeling assumptions that reflect the nature of the task structure, and additionally can result in reduced-rank forms of our model.
In particular, for the hierarchical setting introduced here, we can divide the statistics $\mu$ into \textit{local} and \textit{global} subsets, and for example focus on modeling only the \textit{local} ones to once again reduce to rank-one form.

To motivate with our running example: a finer-grained source that labels \texttt{DOCTOR} versus \texttt{NON-DOCTOR} probably is not accurate on the building type subtask; we can model this source using one accuracy parameter for the former label set (the \textit{local} accuracy) and a different (or no parameter) for the \textit{global} accuracy on irrelevant tasks.
More specifically, for cliques involving $\lf_i$, we can model $P(\lf_i, \y)$ for all $\y$ with only non-$\textit{\texttt{N/A}}$ values in the coverage set of $\lf_i$ using a single parameter, and call this the \textit{local} accuracy; and we can either model $\mu$ for the other $\y$ using one or more other parameters, or simply set it to a fixed value and not model it, to reduce to rank one form, as we do in the experiments.
In particular, this allows us to capture our observation in practice that if a developer is writing a source to distinguish between labels at one sub-tree, they are probably not designing or testing it to be accurate on any of the other subtrees.

  \section{Theoretical Results}
  \label{appendix:theoretical}
\allowdisplaybreaks

In this section, we focus on theoretical results for the basic rank-one model considered in the main body of the paper.
In Section~\ref{appendix:identifiability}, we start by going through the conditions for identifiability in more detail for the rank-one case.
In Section~\ref{appendix:interpreting-bound}, we provide additional interpretation for the expression of our primary theoretical result bounding the estimation error of the label model.
In Section~\ref{appendix:theorem-1}, we then provide the proof of Theorem 1, connecting this estimation error to the generalization error of the end model; and in Section~\ref{appendix:theorem-2}, we provide the full proof of the main bound.

\subsection{Conditions for Identifiability}
\label{appendix:identifiability}
We consider the rank-one setting as in the main body, where we have
\begin{align}
	\label{eqn:rank-one-constraints}
	-(\Sigma_O^{-1})_\Omega
	&=
	\left(zz^T\right)_\Omega,
\end{align}
where $\Omega$ is the inverse augmented edge set, i.e. a pair of indices $(i,j)$, corresponding to elements of $\psi(\mathcal{C})$, and therefore to cliques $A,B \in \mathcal{C}$, is in $\Omega$ if $A,B$ are not part of the same maximal clique in $G_{\text{source}}$ (and therefore $(K_O)_{i,j} = 0$).
This defines a set of $|\Omega|$ equations, which we can encode using a matrix $M_\Omega$, where if $(i,j)$ is the $(r-1)$th entry in $\Omega$, then
\begin{align}
	(M_\Omega)_{r,s}
	&=
	\begin{cases}
		1 & s \in \{i,j\}, \\
		0 & \text{else}.
	\end{cases}
\end{align}
Let $l_i = \log(z_i^2)$ and $q_{(i,j)} = \log(((\Sigma_O^{-1})_{i,j}))$; then by squaring and taking the log of both sides of~\ref{eqn:rank-one-constraints}, we get a system of linear equations:
\begin{align}
	\label{eqn:M-eqn}
	M_\Omega l
	&=
	q_\Omega.
\end{align}
Thus, we can identify $z$ (and therefore $\mu$) \textit{up to sign} if the system of linear equations (\ref{eqn:M-eqn}) has a solution.

\paragraph*{Notes on Invertibility of $M_\Omega$}
Note that if the inverse augmented edge graph consists of a connected triangle (or any odd-numbered cycle), e.g. $\Omega = \{(i,j), (j,k), (i,k)\}$, then we can solve for the $z_i$ up to sign, and therefore $M_\Omega$ must be invertible:
\begin{align*}
	z_i^2
	&=
	\frac{ (\Sigma_O^{-1})_{i,j} (\Sigma_O^{-1})_{i,k} }{ (\Sigma_O^{-1})_{j,k} },
\end{align*}
and so on for $z_j, z_k$.
Note additionally that if other $z_i$ are connected to this triangle, then we can also solve for them up to sign as well.
Therefore, if $\Omega$ contains at least one triangle (or odd-numbered cycle) per connected component, then $M_\Omega$ is invertible.

Also note that this is all in reference to the \textit{inverse} source dependency graph, which will generally be dense (assuming the correlation structure between sources is generally sparse).
For example, note that if we have one source $\lf_i$ that is conditionally independent of all the other sources, then $\Omega$ is fully connected, and therefore if there is a triangle in $\Omega$, then $M_\Omega$ is invertible.

\paragraph*{Identifying the Signs of the $z_i$}
Finally, note that if we know the sign of one $z_i$, then this determines the signs of every other $z_j$ in the same connected component.
Therefore, for $z$ to be uniquely identifiable, we need only know the \textit{sign} of one of the $z_i$ in each connected component.
As noted already, if even one source $\lf_i$ is conditionally independent of all the other sources, then $\Omega$ is fully connected; in this case, we can simply assume that the average source is better than random, and therefore identify the signs of $z$ without any additional information.

\subsection{Interpreting the Main Bound} 
\label{appendix:interpreting-bound}
We re-state Theorem~\ref{thm:mu_est}, which bounds the average error on the estimate of the label model parameters, providing more detail on and interpreting the terms of the bound.

\thmmuest*

\paragraph*{Influence of $\sigma_{\max}(M_{\Omega}^+)$} the largest singular value of the pseudoinverse $M_{\Omega}^+$. Note that $\|M_{\Omega}^+\|^2 = (\lambda_{\min}(M_{\Omega}^TM_{\Omega}))^{-1}$. As we shall see below, $\lambda_{\min}(M_{\Omega}^TM_{\Omega})$ measures a quantity related to the structure of the graph $G_{\text{inv}}$. The smaller this quantity, the more information we have about $G_{\text{inv}}$, and the easier it is to estimate the accuracies. The smallest value of $\|M_{\Omega}^+\|^2$ (corresponding to the largest value of the eigenvalue) is $\sim \frac{1}{\sqrt{m}}$; the square of this quantity in the bound reduces the $m^2$ cost of estimating the covariance matrix to $m$.

It is not hard to see that \[M_{\Omega}^TM_{\Omega}  = \text{diag}(\text{deg}(G_{\text{inv}})) + \text{Adj}(G_{\text{inv}}).\] Here, $\text{deg}(G_{\text{inv}})$ are the degrees of the nodes in $G_{\text{inv}}$ and $\text{Adj}(G_{\text{inv}})$ is its adjacency matrix. This form closely resembles the graph Laplacian, which differs in the sign of the adjacency matrix term: $\mathcal{L}(G) = \text{diag}(\text{deg}(G)) - \text{Adj}(G)$. We bound
\begin{align*}
  \sigma_{\max}(M_\Omega^{+})  \leq  \left(d_{\min} + \lambda_{\min}(\text{Adj}(G_{\text{inv}})))\right)^{-1},
\end{align*}
where $d_{\min}$ is the lowest-degree node in $G_{\text{inv}}$ (that is, the source $s$ with fewest appearances in $\Omega$). In general, computing $ \lambda_{\min}(\text{Adj}(G_{\text{inv}})))$ can be challenging. A closely related task can be done via \emph{Cheeger inequalities}, which state that 
\[2h_G  \geq \lambda_{\min}(\mathcal{L}(G)) \geq \frac{1}{2}h_G^2,\]
where $\lambda_{\min}(\mathcal{L}(G))$ is the smallest non-zero eigenvalue of $\mathcal{L}(G)$ and \[h_G = \min_{X} \frac{|E(X,\bar{X})|}{\min \left \{ \sum_{x \in X} d_x, \sum_{y \in \bar{X}} d_y \right \}}\] is the \emph{Cheeger constant} of the graph \cite{cheeger}. The utility of the Cheeger constant is that it measures the presence of a bottleneck in the graph; the presence of such a bottleneck limits the graph density and is thus beneficial when estimating the structure in our case. Our Cheeger-constant like term $\sigma_{\max}(M_{\Omega}^+)$ acts the same way.

Now, in the easiest and most common case is that of conditionally independent sources~\cite{dalvi:www13,zhang2014spectral,dalvi:www13,karger2011iterative}., $\text{Adj}(G_{\text{inv}})$ has 1's everywhere but the diagonal, and we can compute explicitly that 
\begin{align*}
\sigma_{\max}(M_\Omega^{+})  = \frac{1}{\sqrt{m-2}}.
\end{align*} 
In the general setting, we must compute the minimal eigenvalue of the adjacency matrix, which is tractable, for example, for tree structures.  

\paragraph*{Influence of $\lambda_{\min}(\Sigma_O)$} the smallest eigenvalue of the observed matrix. This quantity reflects the conditioning of the observed (correlation) matrix; the better conditioned the matrix, the easier it is to estimate $\Sigma_O$.

\paragraph*{Influence of $(\Sigma_O^{-1})_{\min}$} the smallest entry of the inverse observed matrix. This quantity contributes to $\Sigma^{-1}$, the geenralized precision matrix that we centrally use; it is a measure of the smallest non-zero correlation between source accuracies (that is, the smallest correlation between non-independent source accuracies). Note that the tail bound of Theorem~\ref{thm:mu_est} scales as $\exp(-((\Sigma_O^{-1})_{\min})^2)$. This is natural, as distinguishing between small correlations and independencies requires more samples.

\subsection{Proof of Theorem 1}
\label{appendix:theorem-1}

Let $\mathcal{D}$ be the true data generating distribution, such that $(\x, \y) \sim \mathcal{D}$.
Let $P_\mu(\y|\lf)$ be the label model parameterized by $\mu$ and conditioned on the observed source labels $\lf$.
Furthermore, assume that:
\begin{enumerate}
	\item For some optimal label model parameters $\mu^*$, $P_{\mu^*}(\lf, \y) = P(\lf, \y)$;
	\item The label $\y$ is independent of the features of our end model given the source labels $\lf$
\end{enumerate}
That is, we assume that (i) the \textit{optimal} label model, parameterized by $\mu^*$, correctly matches the true distribution of source labels $\lf$ drawn from the true distribution, $(s(X), \y) \sim \mathcal{D}$; and (ii) that these labels $\lf$ provide sufficient information to discern the label $\y$.
We note that these assumptions are the ones used in prior work~\citep{ratner2016data}, and are intended primarily to illustrate the connection between the estimation accuracy of $\hat{\mu}$, which we bound in Theorem~\ref{thm:mu_est}, and the end model performance.

Now, suppose that we have an end model parameterized by $w$, and that to learn these parameters we minimize a normalized bounded loss function $l(w, \x, \y)$, such that without loss of generality, $l(w, \x, \y) \leq 1$.
Normally our goal would be to find parameters that minimize the expected loss, which we denote $w^*$:
\begin{align}
	L(w)
	&=
	\E{(\x,\y) \sim \mathcal{D}}{ l(w, \x, \y) }
\end{align}
However, since we do not have access to the true labels $\y$, we instead minimize the expected noise-aware loss, producing an estimate $\tilde{w}$:
\begin{align}
	L_\mu(w)
	&=
	\E{(\x,\y) \sim \mathcal{D}}{ 
		\E{\tilde{\y} \sim P_\mu(\cdot|\lf(\x))}{ l(w, \x, \tilde{\y}) }
	}.
\end{align}
In practice, we actually minimize the \textit{empirical} version of the noise aware loss over an unlabeled dataset $U = \{\x^{(1)},\ldots,\x^{(n)}\}$, producing an estimate $\hat{w}$:
\begin{align}
	\hat{L}_\mu(w)
	&=
	\frac1n \sum_{i=1}^n
		\E{\tilde{\y} \sim P_\mu(\cdot|\lf(\x^{(i)}))}{ 
			l(w, \x^{(i)}, \tilde{\y}) 
		}.
\end{align}
Let $w^*$ be the minimizer of the expected loss $L$, let $\tilde{w}$ be the minimizer of the noise-aware loss for estimated label model parameters $\mu$, $L_\mu$, and let $\hat{w}$ be the minimizer of the empirical noise aware loss $\hat{L}_\mu$.
Our goal is to bound the \textit{generalization risk}- the difference between the expected loss of our empirically estimated parameters and of the optimal parameters,
\begin{align}
	L(\hat{w}) - L(w^*).
\end{align}

Additionally, since analyzing the empirical risk minimization error is standard and not specific to our setting, we simply assume that the error $|L_\mu(\tilde{w}) - L_\mu(\hat{w})| \leq \gamma(n)$, where $\gamma(n)$ is a decreasing function of the number of unlabeled data points $n$.

To start, using the law of total expectation first, followed by our assumption (2) about condtional independence, and finally using our assumption (1) about our optimal label model $\mu^*$, we have that:
\begin{align*}
	L(w)
	&=
	\E{(X', \y') \sim \mathcal{D}}{ L(w) }\\
	&=
	\E{(X', \y') \sim \mathcal{D}}{ 
		\E{(\x,\y) \sim \mathcal{D}}{ l(w, \x', \y) | \x = \x'} 
	}\\
	&=
	\E{(X', \y') \sim \mathcal{D}}{ 
		\E{(\x,\y) \sim \mathcal{D}}{ l(w, \x', \y) | s(\x) = s(\x')} 
	}\\
	&=
	\E{(X', \y') \sim \mathcal{D}}{ 
		\E{(\lf, \tilde{\y}) \sim \mu^*}{
			l(w, \x', \tilde{\y}) | \lf = s(\x')
		} 
	}\\
	&=
	L_{\mu^*}(w).
\end{align*}
Now, we have:
\begin{align*}
	L(\hat{w}) - L(w^*)
	&=
	L_{\mu^*}(\hat{w})
	+ L_\mu(\hat{w}) - L_\mu(\hat{w})
	+ L_\mu(\tilde{w}) - L_\mu(\tilde{w})
	- L_{\mu^*}(w^*) \\
	&\leq
	L_{\mu^*}(\hat{w})
	+ L_\mu(\hat{w}) - L_\mu(\hat{w})
	+ L_\mu(w^*) - L_\mu(\tilde{w})
	- L_{\mu^*}(w^*) \\
	&\leq
	| L_\mu(\hat{w}) - L_\mu(\tilde{w}) |
	+ | L_{\mu^*}(\hat{w}) - L_\mu(\hat{w}) |
	+ | L_\mu(w^*) - L_{\mu^*}(w^*) | \\
	&\leq
	\gamma(n)
	+ 2\max_{w'} | L_{\mu^*}(w') - L_\mu(w') |,
\end{align*}
where in the first step we use our result that $L = L_{\mu^*}$ as well as add and subtract terms; and in the second step we use the fact that $L_\mu(\tilde{w}) \leq L_\mu(w^*)$.
We now have our generalization risk controlled primarily by $| L_{\mu^*}(w') - L_\mu(w') |$, which is the difference between the expected noise aware losses given the estimated label model parameters $\mu$ and the true label model parameters $\mu^*$.
Next, we see that, for any $w'$:
\begin{align*}
	| L_{\mu^*}(w') - L_{\mu}(w') |
	&=
	\left|
		\E{(\x,\y) \sim \mathcal{D}}{ 
			\E{\tilde{\y} \sim P_{\mu^*}(\cdot|\lf)}{ 
				l(w, \x, \tilde{\y}) 
			}
		    - \E{\tilde{\y} \sim P_\mu(\cdot|\lf)}{
		    	l(w, \x, \tilde{\y}) 
		    }
		}
	\right|\\
	&=
	\left|
		\E{(\x,\y) \sim \mathcal{D}}{
			\sum_{\y' \in \mathcal{Y}}
				l(w, \x, \y')
				\left(
					P_{\mu^*}(\y'|\lf) - P_{\mu}(\y'|\lf)
				\right)
		}
	\right|\\
	&\leq
	\sum_{\y' \in \mathcal{Y}}
		\E{(\x,\y) \sim \mathcal{D}}{
			\left| P_{\mu^*}(\y'|\lf) - P_{\mu}(\y'|\lf) \right|
		}\\
	&\leq
	|\mathcal{Y}| \max_{\y'}
		\E{(\x,\y) \sim \mathcal{D}}{
			\left| P_{\mu^*}(\y'|\lf) - P_{\mu}(\y'|\lf) \right|
		},
\end{align*}
where we have now bounded $| L_{\mu^*}(w') - L_{\mu}(w') |$ by the size of the structured output space $|\mathcal{Y}|$, and a term having to do with the difference between the probability distributions of $\mu$ and $\mu^*$.

Now, we use the result from~\citep{honorio2012lipschitz} (Lemma 19) which establishes that the log probabilities of discrete factor graphs with indicator features (such as our model $P_\mu(\lf,\y)$) are $(l_\infty, 2)$-Lipschitz with respect to their parameters, and the fact that for $x,y$ s.t. $|x|, |y| \leq 1$, $|x-y| \leq |\log(x)-\log(y)|$, to get:
\begin{align*}
	\left| P_{\mu^*}(\y'|\lf) - P_{\mu}(\y'|\lf) \right|
	&\leq
	\left| \log P_{\mu^*}(\y'|\lf) - \log P_{\mu}(\y'|\lf) \right|\\
	&\leq
	\left| \log P_{\mu^*}(\lf,\y') - \log P_{\mu}(\lf,\y') \right|
	+ \left|
		\log P_{\mu^*}(\lf)
		- \log P_{\mu}(\lf) \right|\\
	&\leq
	2\norm{ \mu^* - \mu }_\infty + 2\norm{ \mu^* - \mu }_\infty\\
	&\leq
	4\norm{ \mu^* - \mu },
\end{align*}
where we use the fact that the statement of Lemma 19 also holds for every marginal distribution as well.
Therefore, we finally have:
\begin{align*}
	L(\hat{w}) - L(w^*)
	&\leq
	\gamma(n)
	+ 4|\mathcal{Y}| \norm{ \mu^* - \mu }.
\end{align*}

\subsection{Proof of Theorem~\ref{thm:mu_est}}
\label{appendix:theorem-2}
\textit{Proof:} 
First we briefly provide a roadmap of the proof of Theorem~\ref{thm:mu_est}.
We consider estimating $\tilde{\mu}$ with our procedure in the rank-one setting, and we seek a tail bound on $\|\tilde{\mu} - \mu\|$.
The challenge here is that the observed matrix $\Sigma_O$ we see is itself constructed from a series of observed i.i.d. samples $\psi(O)^{(1)}, \ldots, \psi(O)^{(n)}$.
We bound (through a matrix concentration inequality) the error $\Delta_O = \tilde{\Sigma}_O - \Sigma_O$, and view $\Delta_O$ as a perturbation of $\Sigma_O$.
Afterwards, we use a series of perturbation analyses to ultimately bound $\|\tilde{\Sigma}_{O\mathcal{S}} - \Sigma_{O\mathcal{S}} \|$, and then use this directly to bound $\|\tilde{\mu} - \mu\|$; each of the perturbation results is in terms of $\Delta_O$.

We begin with some notation. We write the following perturbations (note that all the terms written with $\Delta$ are additive, while the $\delta$ term is relative)
\begin{align*}
\tilde{\Sigma}_{O\mathcal{S}} = \Sigma_{O\mathcal{S}} + \Delta_{O\mathcal{S}}, \\
\tilde{\Sigma}_O = \Sigma_O + \Delta_O, \\
\tilde{\ell} = \ell + \Delta_{\ell},\\
\tilde{z} = (I + \text{diag}(\delta_z))z.
\end{align*}

Now we start our perturbation analysis:
\begin{align*}
	\tilde{\Sigma}_{O\mathcal{S}}
	&=
	\frac{1}{\sqrt{\tilde{c}}} \tilde{\Sigma}_O \tilde{z}
	=
	\frac{1}{\sqrt{\tilde{c}}} (\Sigma_O + \Delta_O)(I + \text{diag}(\delta_z))z \\
	&=
	\frac{1}{\sqrt{\tilde{c}}} \left(\Sigma_Oz + \Sigma_O\text{diag}(\delta_z)z + \Delta_O(I+\text{diag}(\delta_z))z \right).
\end{align*}

Subtracting $\Sigma_{O\mathcal{S}} = \frac{1}{\sqrt{c}} \Sigma_O z$, we get
\begin{equation}
	\Delta_{O\mathcal{S}}
	=
	\left(\frac{1}{\sqrt{\tilde{c}}}-\frac{1}{\sqrt{c}} \right) \Sigma_O z
	+ \frac{1}{\sqrt{\tilde{c}}} \left( \Sigma_O\text{diag}(\delta_z)z
	+ \Delta_O(I+\text{diag}(\delta_z))z \right).
	\label{eq:maindmu}
\end{equation}

The rest of the analysis requires us to bound the norms for each of these terms. 

{\bf Left-most term}. We have that
\begin{align*}
	\left\|\left( \frac{1}{\sqrt{\tilde{c}}} - \frac{1}{\sqrt{c}} \right)\Sigma_Oz \right\|
	&=
	\left| \frac{\sqrt{c}}{\sqrt{\tilde{c}}} -1 \right| \left\| \frac{1}{\sqrt{c}} \Sigma_Oz\right\|
	= 
	\left| \frac{\sqrt{c}}{\sqrt{\tilde{c}}} -1\right| \|\Sigma_{O\mathcal{S}}\|
	\leq
	\sqrt{d_O} \left| \frac{\sqrt{c}}{\sqrt{\tilde{c}}} -1 \right | \leq
	\sqrt{d_O} |\tilde{c} - c|.
\end{align*}

Here, we bounded $\|\Sigma_{O\mathcal{S}}\|$ by $\sqrt{d_O}$, since $\Sigma_{O\mathcal{S}} \in [-1,1]^{d_O}$. 
Then, note that $c = \Sigma_\mathcal{S}^{-1} (1 + z^T\Sigma_O z) \geq 0$, since $\Sigma_\mathcal{S} < 1$ and $\Sigma_O \succeq 0 \implies z^T\Sigma_O z \geq 0$, so therefore $c, \tilde{c} \geq 1$.
In the last inequality, we use this to imply that $|\sqrt{c}/\sqrt{\tilde{c}}-1| \leq |\sqrt{c}-\sqrt{\tilde{c}}| \leq |\tilde{c} - c|$. 
Next we work on bounding $|\tilde{c} - c|$. We have
\begin{align*}
	|\tilde{c}-c|
	&=
	|\Sigma_\mathcal{S}^{-1}||\tilde{z}^T\tilde{\Sigma}_O\tilde{z} - z^T\Sigma_Oz| \\
	&=
	|\Sigma_\mathcal{S}^{-1}||z^T (I+\text{diag}(\delta_z))^T(\Sigma_O + \Delta_O)(I+\text{diag}(\delta_z))z - z^T\Sigma_Oz|\\
	&=
	|\Sigma_\mathcal{S}^{-1}||z^T\Sigma_O\text{diag}(\delta_z)z + z^T\Delta_O(I+\text{diag}(\delta_z))z + z^T \text{diag}(\delta_z)^T (\Sigma_O+\Delta_O) (I+\text{diag}(\delta_z)) z | \\
	&\leq
	|\Sigma_\mathcal{S}^{-1}|\|z\|^2 \left( 
		\|\Sigma_O\| \left( 2\|\delta_z\| + \|\delta_z\|^2 \right)
		+ \|\Delta_O\| \left( 2\|\delta_z\| + \|\delta_z\|^2 + 1\right)
	\right)\\
	&\leq
	\|z\|^2 \left( 
		\|\Sigma_O\| \left( 2\|\delta_z\| + \|\delta_z\|^2 \right)
		+ \|\Delta_O\| \left( 2\|\delta_z\| + \|\delta_z\|^2 + 1\right)
	\right).
\end{align*}

Thus,
\begin{align}
	\left\|\left( \frac{1}{\sqrt{\tilde{c}}} - \frac{1}{\sqrt{c}} \right)\Sigma_Oz \right\|
	&\leq
	\sqrt{d_O}\|z\|^2 \left( 
		\|\Sigma_O\| \left( 2\|\delta_z\| + \|\delta_z\|^2 \right)
		+ \|\Delta_O\| \left( 2\|\delta_z\| + \|\delta_z\|^2 + 1\right)
	\right).
	\label{eq:cterm}
\end{align}

{\bf Bounding $c$}. We will need a bound on $c$ to bound $z$. We have that 
	\[ c = (\Sigma_\mathcal{S}-\Sigma_{O\mathcal{S}}^T \Sigma_O^{-1} \Sigma_{O\mathcal{S}})^{-1}.\]
Applying the Woodbury matrix inversion lemma, we have:
\begin{align*}
	c
	&=
	\Sigma_{\mathcal{S}}^{-1}
	+
	\Sigma_{\mathcal{S}}^{-1} \Sigma_{O\mathcal{S}}^T \left(
		\Sigma_{O} - \Sigma_{O\mathcal{S}}\Sigma_{\mathcal{S}}^{-1}\Sigma_{O\mathcal{S}}^T
	\right)^{-1}
	\Sigma_{O\mathcal{S}}\Sigma_{\mathcal{S}}^{-1}
\end{align*}
Now, by the blockwise inversion lemma, we know that
\begin{align*}
	K_O
	&=
	\left(
		\Sigma_{O} - \Sigma_{O\mathcal{S}}\Sigma_{\mathcal{S}}^{-1}\Sigma_{O\mathcal{S}}^T
	\right)^{-1}
\end{align*}
So we then have:
\begin{align*}
	c
	&=
	\Sigma_{\mathcal{S}}^{-1}
	+
	\Sigma_{\mathcal{S}}^{-1} \Sigma_{O\mathcal{S}}^T K_O \Sigma_{O\mathcal{S}}\Sigma_{\mathcal{S}}^{-1}
	\leq
	\Sigma_{\mathcal{S}}^{-1}
	+
	(\Sigma_{\mathcal{S}}^{-1})^2 \|\Sigma_{O\mathcal{S}}\|^2 \|K_O\|
\end{align*}

{\bf Bounding $z$}. We'll  use our bound on $c$, since $z = \sqrt{c}\Sigma_O^{-1}\Sigma_{O\mathcal{S}}$. 
\begin{align*}
	\|z\|
	&=
	\|\sqrt{c}\Sigma_O^{-1}\Sigma_{O\mathcal{S}}\| \\
	&\leq 
	\left( 
		\Sigma_{\mathcal{S}}^{-1}
		+ (\Sigma_{\mathcal{S}}^{-1})^2 \|\Sigma_{O\mathcal{S}}\|^2 \|K_O\|
	\right)^{\frac12}
	\|\Sigma_O^{-1}\| \|\Sigma_{O\mathcal{S}}\|\\
	&\leq 
	\left( 
		\Sigma_{\mathcal{S}}^{-1}
		+ (\Sigma_{\mathcal{S}}^{-1})^2 d_O \|K_O\|
	\right)^{\frac12}
	\|\Sigma_O^{-1}\| \sqrt{d_O}\\
	&=
	\frac{d_O}{\Sigma_{\mathcal{S}}} \left( 
		\frac{\Sigma_{\mathcal{S}}}{d_O}
		+ \lambda_{\textrm{max}}(K_O)
	\right)^{\frac12}
	\lambda_{\textrm{min}}^{-1}(\Sigma_O)
\end{align*}

In the last inequality, we used the fact that $\|\Sigma_{O\mathcal{S}}\|^2 \leq d_O$. Now we want to control $\|\Delta_{\ell}\|$.  

{\bf Perturbation bound}. We have the perturbation bound
\begin{equation}
 \|\Delta_{\ell}\| \leq \|M_\Omega^+\| \|\tilde{q}_S - q_S\|.
\label{eq:lpb}
 \end{equation}

We need to work on the term $\|\tilde{q}_S - q_S \|$. To avoid overly heavy notation, we write $P = \Sigma_O^{-1}$, $\tilde{P} = \tilde{\Sigma}_O^{-1}$, and $\Delta_P = P - \tilde{P}$. 
Then we have:
\begin{align*}
	\|\tilde{q}_S-q_S\|^2
	&=
	\sum_{(i,j)\in S} \left( \log(\tilde{P}_{i,j}^2) - \log(P_{i,j}^2) \right)^2 \\
	&=
	4 \sum_{(i,j)\in S} \left( \log(|\tilde{P}_{i,j}|) - \log(|P_{i,j}|) \right)^2 \\
	&= 
	4 \sum_{(i,j)\in S} \left( \log(|P_{i,j} + (\Delta_P)_{i,j}|) - \log(|P_{i,j}|) \right)^2 \\
	&\leq
	4 \sum_{(i,j)\in S} \left[ \log \left(1 + \left|\frac{(\Delta_P)_{i,j}}{P_{i,j}} \right| \right)\right]^2  \\
	&\leq 
	8 \sum_{(i,j) \in S} \left( \frac{|(\Delta_P)_{i,j}|}{|P_{i,j}|} \right)^2  \\
	&\leq 
	\frac{8}{P^2_{\min}} \sum_{(i,j) \in S} (\Delta_P)_{i,j}^2 \\
	&\leq
	\frac{8 \|\tilde{\Sigma}_O^{-1}-\Sigma_O^{-1}\|^2}{((\Sigma_O^{-1})_{\min})^2}.
\end{align*}
Here, the second inequality uses $(\log(1+x))^2 \leq x^2$, and the fourth inequality sums over squared values. Next, we use the perturbation bound $\|\tilde{\Sigma}_O^{-1} - \Sigma_O^{-1}\| \leq \|\Sigma_O^{-1}\|^2 \|\Delta_O\|$, so that we have 
\begin{align*}
\|\tilde{q}_S-q_S\| \leq\frac{2\sqrt{2} \|\Sigma_O^{-1}\|^2\|\Delta_O\|}{(\Sigma_O^{-1})_{\min}}.
\end{align*}

Then, plugging this into \eqref{eq:lpb}, we get that

\begin{equation}
	\|\Delta_{\ell}\|
	\leq
	\sigma_{\max}(M_\Omega^+)
	\frac{ 2\sqrt{2}\|\Sigma_O^{-1}\|^2 \|\Delta_O\|}{(\Sigma_O^{-1})_{\min}}.
\label{eq:ellbound}
\end{equation}

{\bf Bounding} $\delta_z$. Note also that $\|\Delta_{\ell}\|^2 = \sum_{i=1}^m (\log(\tilde{z}_i^2)-\log(z_i^2))$. We have that
\begin{align*}
\|\Delta_{\ell}\|^2 &= \sum_{i=1}^m \log \left( \frac{\tilde{z}_i^2}{z_i^2} \right) \\
&=2 \sum_{i=1}^m \log \left( \frac{|\tilde{z}_i|}{|z_i|} \right) \\
&= 2 \sum_{i=1}^m \log (1+|(\delta_z)_i|),\\
& \geq 2 \sum_{i=1}^m (\delta_z)_i^2 \\
&= 2\|\delta_z\|^2,
\end{align*}
where in the fourth step, we used the bound $\log(1+a) \geq a^2$ for small $a$. Then, we have
\begin{equation}
\|\delta_z\| \leq \frac{ \sqrt{2}\|\Sigma_O^{-1}\|^2 \|\Delta_O\|}{(\Sigma_O^{-1})_{\min}} \sigma_{\max}(M_\Omega^+).
\label{eq:dzbound}
\end{equation}
 
{\bf Putting it together}.
Using \eqref{eq:maindmu}, we have that
\begin{align*}
	\|\Delta_{O\mathcal{S}} \|
	&=
	\left \| \left(\frac{1}{\sqrt{\tilde{c}}}-\frac{1}{\sqrt{c}} \right)\Sigma_Oz +  \frac{1}{\sqrt{\tilde{c}}} \left( \Sigma_O\text{diag}(\delta_z)z + \Delta_O(I+\text{diag}(\delta_z))z \right) \right\| \\
	&\leq 
	\left\|  \left(\frac{1}{\sqrt{\tilde{c}}}-\frac{1}{\sqrt{c}} \right)\Sigma_Oz \right \| + \left( \| \Sigma_O\text{diag}(\delta_z) \| + \|\Delta_O(I+\text{diag}(\delta_z))  \| \right) \|z\| \\
	&\leq
	\sqrt{d_O}\|z\|^2 \left( 
		\|\Sigma_O\| \left( 2\|\delta_z\| + \|\delta_z\|^2 \right)
		+ \|\Delta_O\| \left( 2\|\delta_z\| + \|\delta_z\|^2 + 1\right)
	\right)\\
	&\qquad \qquad
	+ \| \Sigma_O \| \|\delta_z\| \|z\|
	+ \|\Delta_O\| \|z\| (1 + \|\delta_z\|)\\
	&\leq
	\sqrt{d_O}\|z\|^2 \left( 
		3 \|\Sigma_O\| \|\delta_z\|
		+ 3 \|\Delta_O\| \|\delta_z\|
		+ \|\Delta_O\|
	\right)\\
	&\qquad \qquad
	+ \| \Sigma_O \| \|\delta_z\| \|z\|
	+ \|\Delta_O\| \|z\| (1 + \|\delta_z\|)\\
	&\leq
	\|z\| \left( 3\sqrt{d_O}\|z\| + 1 \right)
	\left(
		(\|\Sigma_O\| + \|\Delta_O\|) \|\delta_z\|
		+ \|\Delta_O\|
	\right)
\end{align*}
Where in the first inequality, we use the triangle inequality and the fact that $\tilde{c} > 1$, and in the third inequality, we relied on the fact that we can control $\|\delta_z\|$ (through $\|\Delta_O\|$) so that we can make it small enough and thus take $\|\delta_z\|^2 \leq \|\delta_z\|$.
Now we can plug in our bounds on $\|z\|$ and $\|\delta_z\|$ from before:
\begin{align*}
	\|\Delta_{O\mathcal{S}} \|
	&\leq
	\left(
		\frac{d_O}{\Sigma_{\mathcal{S}}} \left( 
			\frac{\Sigma_{\mathcal{S}}}{d_O}
			+ \lambda_{\textrm{max}}(K_O)
		\right)^{\frac12}
		\lambda_{\textrm{min}}^{-1}(\Sigma_O)
	\right)
	\left(
		3\sqrt{d_O}\left(
			\frac{d_O}{\Sigma_{\mathcal{S}}} \left( 
				\frac{\Sigma_{\mathcal{S}}}{d_O}
				+ \lambda_{\textrm{max}}(K_O)
			\right)^{\frac12}
			\lambda_{\textrm{min}}^{-1}(\Sigma_O)
		\right)
		+ 1
	\right)\\
	&\times
	\left(
		(\|\Sigma_O\| + \|\Delta_O\|) \left( \frac{ \sqrt{2}\|\Sigma_O^{-1}\|^2 \|\Delta_O\|}{(\Sigma_O^{-1})_{\min}} \sigma_{\max}(M_\Omega^+) \right)
		+ \|\Delta_O\|
	\right)
\end{align*}

For convenience, we set $\|\Delta_O\| = t$.
Recall that
\begin{align*}
	a
	&=
	\left( 
		\frac{d_O}{\Sigma_{\mathcal{S}}} 
		+ \left(\frac{d_O}{\Sigma_{\mathcal{S}}}\right)^2 \lambda_{\textrm{max}}(K_O)
	\right)^{\frac12}
\end{align*}
and
\begin{align*}
	b
	&=
	\frac{ \|\Sigma_O^{-1}\|^2}{(\Sigma_O^{-1})_{\min}}.
\end{align*}
Then, we have
 \[ \|\Delta_{O\mathcal{S}} \| \leq (3\sqrt{d_O}a \lambda_{\min}^{-1}(\Sigma_O)+1)\left(\sqrt{2}ab \kappa(\Sigma_O)  \sigma_{\max}(M_\Omega^+) t + \sqrt{2}ab \frac{ \sigma_{\max}(M_\Omega^+)}{  \lambda_{\min}(\Sigma_O)}t^2 + a   \lambda_{\min}^{-1}(\Sigma_O)t \right).\]

Again we can take $t$ small so that $t^2 \leq t$. Simplifying further, we have
\[ \|\Delta_{O\mathcal{S}} \| \leq (3\sqrt{d_O}a \lambda_{\min}^{-1}(\Sigma_O)+1)\left(\sqrt{2}ab \sigma_{\max}(M_\Omega^+) \left[ \kappa(\Sigma_O)   +\lambda_{\min}^{-1}(\Sigma_O)\right] + a   \lambda_{\min}^{-1}(\Sigma_O) \right)t.\]
Finally, since the $a\lambda_{\min}^{-1}(\Sigma_O)$ is smaller than the left-hand term inside the parentheses, we can write
\begin{equation}
\label{eq:tbound}
\|\Delta_{O\mathcal{S}} \| \leq (3\sqrt{d_O}a \lambda_{\min}^{-1}(\Sigma_O)+1)\left(2\sqrt{2}ab \sigma_{\max}(M_\Omega^+) \left[ \kappa(\Sigma_O)   +\lambda_{\min}^{-1}(\Sigma_O)\right] \right)t.
\end{equation}
 
{\bf Concentration bound.}
We need to bound $t = \|\Delta_O\|$, the error when estimating $\Sigma_O$ from observations $\psi(O)^{(1)}, \ldots, \psi(O)^{(n)}$  over $n$ unlabeled data points.

To start, recall that $O$ is the set of observable cliques, $\psi(O) \in \{0,1\}^{d_O}$ is the corresponding vector of minimal statistics, and $\Sigma_O = \Cov{}{ \psi(O) }$.
For notational convenience, let $R = \E{}{ \psi(O)\psi(O)^T }$, $r = \E{}{ \psi(O) }$, and $r_k = \psi(O)^{(k)}$, and $\Delta_r = \frac1n\sum_{i=1}^n r_k - r$.
Then we have:
\begin{align*}
	\norm{ \Delta_O }
	=
	\norm{ \Sigma_O - \tilde{\Sigma}_O }
	&=
	\norm{ 
		(R - rr^T) 
		- \left( 
			\frac1n \sum_{i=1}^n r_ir_i^T 
			- \left( r + \Delta_r \right)
				\left( r + \Delta_r \right)^T
		\right)
	}\\
	&\leq
	\underbrace{ \norm{ R - \frac1n \sum_{i=1}^n r_ir_i^T } }_{\Delta_R}
	+ \underbrace{ \norm{ 
		rr^T - \left( r + \Delta_r \right) \left( r + \Delta_r \right)^T 
	} }_{\Delta_r}.
\end{align*}
We start by applying the matrix Hoeffding inequality~\citep{tropp2015introduction} to bound the first term, $\Delta_R$.
Let $S_k = \frac1n (R - R_k)$, and thus clearly $\E{}{S_k} = 0$.
We seek a sequence of symmetric matrices $A_k$ s.t. $S_k^2 \preceq A_k^2$.
First, note that, for some vectors $x,v$,
\begin{align*}
	x^T\left( \norm{ v }^2I - vv^T \right)x
	&=
	\norm{ v }^2 \norm{ x }^2 - \ip{x}{v}^2
	\geq
	0
\end{align*}
using Cauchy-Schwarz; therefore $\norm{v}^2I \succeq vv^T$, so that
\begin{align*}
	d_O^2I 
	&\succeq
	\norm{ r_k }^4I
	\succeq
	\norm{ r_k }^2 r_kr_k^T 
	=
	(r_kr_k^T)^2.
\end{align*}
Next, note that $(r_kr_k^T + R)^2 \succeq 0$.
Now, we use this to see that:
\begin{align*}
	(nS_k)^2
	&=
	(r_kr_k^T - R)^2
	\preceq
	(r_kr_k^T - R)^2 + (r_kr_k^T + R)^2
	=
	2((r_kr_k^T)^2 + R^2)
	\preceq
	2(d_O^2I + R^2).
\end{align*}
Therefore, let $A_k^2 = \frac{2}{n^2}(d_O^2I + R^2)$, and note that $\norm{R^2} \leq \norm{R}^2 \leq (d_O\norm{R}_{max})^2 = d_O^2$.
We then have
\begin{align*}
	\sigma^2
	&=
	\norm{ \sum_{k=1}^n A_k^2 }
	\leq
	\frac{2}{n}\left( d_O^2 + \norm{R^2} \right)
	\leq
	\frac{4d_O^2}{n}.
\end{align*}
And thus,
\begin{align}
	P\left( \norm{ \Delta_R } \geq \gamma \right)
	&\leq
	2d_O\exp\left(
		-\frac{
			n\gamma^2
		}{
			32d_O^2
		}
	\right).
	\label{eqn:matrix_hoeffding}
\end{align}

Next, we bound $\Delta_r$.
We see that:
\begin{align*}
	\norm{ \Delta_r }
	&=
	\norm{ 
		rr^T 
		- \left( r + \Delta_r \right) \left( r + \Delta_r \right)^T 
	}\\
	&=
	\norm{ 
		r\Delta_r^T + \Delta_rr^T + \Delta_r\Delta_r^T
	}\\
	&\leq
	\norm{ r\Delta_r^T } 
	+ \norm{ \Delta_rr^T } 
	+ \norm{ \Delta_r\Delta_r^T }\\
	&\leq
	2\norm{r} \norm{\Delta_r}
	+ \norm{ \Delta_r }^2\\
	&\leq
	3 \norm{r} \norm{\Delta_r}\\
	&\leq
	3 \norm{r}_1 \norm{\Delta_r}_1\\
	&\leq
	3 d_O^2 | \Delta_r' |,
\end{align*}
where $\Delta_r'$ is the perturbation for a single element of $\psi(O)$.
We can then apply the standard Hoeffding's bound to get:
\begin{align*}
	P(\norm{ \Delta_r } \geq \gamma)
	&\leq
	2\exp\left( -\frac{ 2n\gamma^2 }{3d_O^2} \right),
\end{align*}
Combining the bounds for $\norm{\Delta_R}$ and $\norm{\Delta_r}$, we get:
\begin{align}
	P(\|\Delta_O\| \geq \gamma)
	&=
	P(t \geq \gamma)
	\leq
	3d_O \exp \left( -\frac{n\gamma^2}{32d_O^2} \right).
\end{align}
 
{\bf Final steps} 
Now, we use the bound on $t$ in \eqref{eq:tbound} and the concentration bound above to write
 \begin{align*} 
 P(\|\Delta_{O\mathcal{S}}\| \geq t') &\leq P(Vt \geq t') \\
 &= P\left(t \geq \frac{t'}{V} \right) \\
& \leq  2d_O \exp \left( -\frac{nt'^2}{32V^2d_O^2} \right),
 \end{align*}
 where $V= (3\sqrt{d_O}a \lambda_{\min}^{-1}(\Sigma_O)+1)\left(2\sqrt{2}ab \sigma_{\max}(M_\Omega^+) \left[ \kappa(\Sigma_O)   +\frac{1}{  \lambda_{\min}(\Sigma_O)}\right] \right)$.

Given $\tilde{\Sigma}_{O\mathcal{S}}$, we recover $\tilde{\mu}_1 = \tilde{\Sigma}_{O\mathcal{S}} + \E{}{\psi(H)}\Ehat{\psi(O)}$.
We assume $\E{}{\psi(H)}$ is known, and we can bound the error introduced by $\E{}{\psi(H)}\Ehat{\psi(O)}$ as above, which we see can be folded into the looser bound for the error in $\tilde{\Sigma}_{O\mathcal{S}}$.

Finally, we expand the rank-one form $\tilde{\mu}_1$ into $\tilde{\mu}$ algebraically, according to our weight tying in the rank one model we use.
Suppose in the rank one reduction (see Section~\ref{appendix:rank-one-reduction}), we let $\y_B = \ind{ \y = y_1 }$.
Then each element of $\mu_1$ that we track corresponds to either the probability of being correct, $\alpha_{C,y} = P(\cap_{i\in C}\{\lf_i = y\}, \y=y)$ or the probability of being incorrect, $\frac{1}{r-1}(1-\alpha_{C,y})$, for each source clique $C$ and label output combination $y_C$, and this value is simply copied $r-1$ times (for the other, weight-tied incorrect values), except for potentially one entry where it is multiplied by $(r-1)$ and then subtracted from $1$ (to transform from incorrect to correct).
Therefore, $\norm{\Delta_\mu} = \norm{ \mu - \tilde{\mu} } \leq 2(r-1) \norm{ \mu_1 - \tilde{\mu}_1 }$.
Thus, we have:
\begin{align*} 
	P(\|\Delta_{\mu}\| \geq t')
	&\leq
	4(r-1)d_O \exp \left( -\frac{nt'^2}{32V^2d_O^2} \right),
\end{align*}
where $V$ is defined as above.
We only have one more step:
\begin{align*}
	\E{}{\norm{ \tilde{\mu} - \mu}}
	&=
	\int_{0}^{\infty} P(\|\tilde{\mu} - \mu\| \geq \gamma) d\gamma \\
	&\leq
	\int_{0}^{\infty} 4(r-1)d_O \exp \left( -\frac{n}{32V^2d_O^2} \gamma^2 \right)d\gamma \\
	&=
	\frac{4(r-1)d_O \sqrt{\pi}}{2 \sqrt{\frac{n}{32V^2d_O^2}}} \\
	&=
	4(r-1)d_O^2 \sqrt{\frac{32 \pi}{n}}V.
\end{align*}
Here, we used the fact that $\int_0^{\infty} \exp(-a\gamma^2) d\gamma = \frac{\sqrt{\pi}}{2 \sqrt{a}}$. \hfill $\square$

  \section{Experimental Details}
  \label{appendix:exp_details}

\subsection{Data Balancing and Label Model Training Procedure}
For each application, rebalancing was applied via direct subsampling to the training set in the manner that was found to most improve development set micro-averaged accuracy.  Specifically, we rebalance with respect to the median class for OpenI (i.e., removing examples from majority class such that none had more than the original median class), the minimum class for TACRED, and perform no rebalancing for OntoNotes.  For generative model training, we use stochastic gradient descent with a step size, step number, and $\ell_2$ penalty listed in Table \ref{table:hyperparameters} below.  These parameters were found via 10-trial coarse random search, with all values determined via maximum micro-averaged accuracy evaluated on the development set. 

\begin{table}[h!]
    \begin{center}
    \begin{tabular}{lrrr} 
        \toprule
        & OntoNotes & TACRED  & OpenI  \\ \midrule \midrule
        \textbf{Label Model Training} & & \\ \midrule
        Step Size & 5e-3 & 1e-2 & 5e-4 \\
        $\ell_2$  Regularization& 1e-4 &  4e-4 & 1e-3 \\
        Step Number &  50 & 25 & 50\\
        \midrule
        \textbf{End Model Architecture} & & \\ \midrule
        Embedding Initialization & PubMed & FastText EN & Random \\
        Embedding Size & 100 & 300 & 200\\
        LSTM Hidden Size & 150 & 250 & 150 \\
        LSTM Layers & 1 & 2 & 1 \\
        Intermediate Layer Dimensions & 200, 50 & 200, 50, 25  & 200, 50\\ \midrule
        \textbf{End Model Training} & & \\ \midrule
        Learning Rate & 1e-2 & 1e-3 & 1e-3 \\
        $\ell_2$ Regularization & 1e-4 & 1e-4  & 1e-3 \\
        Epochs & 20 & 30 & 50 \\
        Dropout & 0.25 & 0.25 & 0.1 \\ 
        \bottomrule
    \end{tabular}
    \caption{Model architecture and training parameter details.}
    \label{table:hyperparameters}
    \end{center}
    \end{table}

\subsection{End Model Training Procedure}
Before training over multiple iterations to attain averaged results for reporting, a 10-trial random search over learning rate and $\ell_2$ regularization with the Adam optimizer was performed for each application based on micro-averaged development set accuracy.  Learning rate was decayed by an order of magnitude if no increases in training loss improvement or development set accuracy were observed for 10 epochs, and the learning rate was frozen during the first 5 epochs.  Models are reported using early stopping, wherein the best performing model on the development set is eventually used for evaluation on the held-out test set, and maximum epoch number is set for each application at a point beyond which minimal additional decrease in training loss was observed.

\subsection{Dataset Statistics}
We give additional detail in here (see Table \ref{tab:datasets}) on the different datasets used for the experimental portion of this work.  All data in the development and test sets is labeled with ground truth, while data in the training set is treated as unlabeled.  Each dataset has a particular advantage in our study.  The OntoNotes set, for instance, contains a particularly large number of relevant data points (over 63k), which enables us to investigate empirical performance scaling with the number of unlabeled data points.  Further, the richness of the TACRED dataset allowed for the creation of an 8-class, 7-sub-task hierarchical classification problem, which demonstrates the utility of being able to supervise at each of the three levels of task granularity.  Finally, the OpenI dataset represents a real-world, non-benchmark problem drawn from the domain of medical triage, and domain expert input was directly leveraged to create the relevant supervision sources.  The fact that these domain expert weak supervision sources naturally occurred at multiple levels of granularity, and that the could be easily integrated to train an effective end model, demonstrates the utility of the \systemx framework in practical settings.
\begin{table}[hbt!]
\setlength\tabcolsep{5pt}
{
\vspace{10pt}
\centering
\begin{tabular}{lrrrccccccc}
      \toprule
           & \# Train & \# Dev & \# Test & Tree Depth & \# Tasks & \# Sources/Task \\
      \midrule
      OntoNotes (NER)
        & 62,547
        & 350
        & 345
        & 2
        & 3
        & 11
        \\
      TACRED (RE)
        & 9,090
        & 350
        & 2174
        & 3
        & 7
        & 9
        \\
      OpenI (Doc)
        & 2,630
        & 200
        & 378
        & 2
        & 3
        & 19
        \\
      \bottomrule\\
    \end{tabular}
    \caption{Dataset split sizes and sub-task structure for the three fine-grained classification tasks on which we evaluate \systemx.}
    \label{tab:datasets}
}
\end{table}

\subsection{Task Accuracies}
\label{appendix:task_accuracies}

For clarity, we present in Table \ref{table:task_accuracies} the individual task accuracies of both the learned \systemx model and MV for each experiment.  These accuracies are computed from the output of evaluating each model on the test set with ties broken randomly.  
      
\begin{table}[hbt!]
\begin{center}
\begin{tabular}{cccc} \toprule
   & OntoNotes  & TACRED  & OpenI  \\ \midrule    
  \underline{Task 1}  & & &\\
  MV   & 93.3  & 74.2  & 83.9 \\
  \systemx & 91.9  & 80.5 & 84.1 \\ \midrule
    \underline{Task 2}  & & &\\
  MV   & 73.3  & 46.2  &  77.8 \\
  \systemx & 75.6  & 65.9 & 83.7 \\ \midrule
      \underline{Task 3}  & & &\\
  MV   & 71.4  & 74.9 & 61.7 \\
  \systemx & 74.1   & 74.8 & 61.7 \\ \midrule
        \underline{Task 4}  & & &\\
  MV   & -  & 34.4 & - \\
  \systemx &  - & 60.2 & - \\ \midrule
          \underline{Task 5}  & & &\\
  MV   & -  & 36.2 & - \\
  \systemx &  - & 40.2 & - \\ \midrule
            \underline{Task 6}  & & &\\
  MV   & -  & 56.3 & - \\
  \systemx &  - &  49.9 & - \\ \midrule
              \underline{Task 6}  & & &\\
  MV   & -  & 36.8 & - \\
  \systemx &  - &  56.3 & - \\ 
  
\bottomrule
\end{tabular}
\caption{Label model task accuracies for each task for for both our approach and majority vote (\systemx/MV)}
\label{table:task_accuracies}
\end{center}
\end{table}

\subsection{Ablation Study: Unipolar Correction and Joint Modeling}
\label{appendix:unipolar}

We perform an additional ablation to demonstrate the relative gains of modeling unipolar supervision sources and jointly modeling accuracies across multiple tasks with respect to the data programming (DP) baseline \citep{ratner2018snorkel}.  Results of this investigation are presented in Table \ref{tab:unipolar}. We observe an average improvement of $\AvgUnipolarBoost$ points using the unipolar correction (DP-UI), and an additional 1.3 points from joint modeling within \systemx, resulting in an aggregate gain of $\AvgGainOverDP$  accuracy points over the data programming baseline.

\begin{table}[hbt!]
    \centering
    \begin{tabular}{lrrrr}
      \toprule
              & OntoNotes (NER) & TACRED (RE) & OpenI (Doc) & Average\\
      \midrule
      DP \citep{ratner2016data}
        & 78.4 $\pm$ 1.2
        & 49.0 $\pm$ 2.7
        & 75.8 $\pm$ 0.9
        & 67.7
        \\
       \midrule
      DP-UI
         & 81.0 $\pm$ 1.2
         & 54.2 $\pm$ 2.6
         & 76.4 $\pm$ 0.5
         & {70.5}
         \\
      \midrule
      \systemx
        & \textbf{82.2} $\pm$ 0.8
        & \textbf{56.7} $\pm$ 2.1
        & \textbf{76.6} $\pm$ 0.4
        & \textbf{71.8}
        \\
      \bottomrule
    \end{tabular}
    \caption{\textbf{Effect of Unipolar Correction.} We compare the micro accuracy (avg. over 10 trials) with 95\% confidence intervals of a model trained using data programming (DP), data program with a unipolar correction (DP-UI), and our approach (\systemx).}
    \label{tab:unipolar}
\end{table}

\end{appendix}

\end{document}